\documentclass[11pt,a4paper]{article}

\usepackage[utf8]{inputenc}
\usepackage[T1]{fontenc}
\usepackage{amsmath,amssymb,amsthm}
\usepackage{graphicx}
\usepackage{hyperref}
\usepackage{cleveref}
\usepackage[normalem]{ulem}
\usepackage{booktabs}

\usepackage{ltablex}    
\usepackage{multirow}
\usepackage{csvsimple}
\usepackage{float}
\usepackage[margin=1in]{geometry}
\usepackage{changepage} 
\usepackage{caption}

\usepackage{newpxtext}
\usepackage{newpxmath}

\newcommand{\hitomi}{\textit{HitoMi-Cam}}

\newlength{\extralength}
\newlength{\fulllength}
\title{HitoMi-Cam: A Shape-Agnostic Person Detection Method Using the Spectral Characteristics of Clothing}

\author{Shuji Ono\\
\small Fujifilm Corporation, Kaisei, Ashigara-kami, Kanagawa 258-8577, Japan\\
\small \texttt{shuji.ono@fujifilm.com, shuji.ono.1960@gmail.com}\\
\small ORCID: 0000-0001-9639-7863}

\date{Submitted to Journal of Imaging (under review)}

\begin{document}

\maketitle

\begin{abstract}
While convolutional neural network (CNN)-based object detection is widely used, it exhibits a shape dependency that degrades performance for postures not included in the training data. Building upon our previous simulation study published in Journal of Imaging~\cite{ono2025jimaging}, this study implements and evaluates the spectral-based approach on physical hardware to address this limitation. Specifically, this paper introduces HitoMi-Cam, a lightweight and shape-agnostic person detection method that uses the spectral reflectance properties of clothing. The author implemented the system on a resource-constrained edge device without a GPU to assess its practical viability. The results indicate that a processing speed of 23.2 frames per second (fps) (253 × 190 pixels) is achievable, suggesting that the method can be used for real-time applications. In a simulated search and rescue scenario where the performance of CNNs declines, HitoMi-Cam achieved an average precision (AP) of 93.5\%, surpassing that of the compared CNN models (best AP of 53.8\%). Throughout all evaluation scenarios, the occurrence of false positives remained minimal. This study positions the HitoMi-Cam method not as a replacement for CNN-based detectors but as a complementary tool under specific conditions. The results indicate that spectral-based person detection can be a viable option for real-time operation on edge devices in real-world environments where shapes are unpredictable, such as disaster rescue.
\end{abstract}


\section{Introduction}

In recent years, convolutional neural network (CNN)-based object detectors, typified by the You Only Look Once (YOLO) system, have become a standard technology in various applications such as autonomous driving, surveillance cameras, and robotics, due to their high accuracy and real-time processing capabilities \cite{redmon2016yolo, terven2023yolo_review}.
These methods achieve high performance by learning spatial intensity patterns from large-scale image datasets to recognize the shape and texture of target objects \cite{girshick2014rich}.

However, such CNN-based methods face a challenge of insufficient robustness stemming from statistical biases in the training datasets \cite{torralba2011unbiased, tommasi2017deeper}.
For instance, in the MS COCO dataset \cite{lin2014microsoft}, a major benchmark for object detection, approximately 85\% of person instances are in an upright posture, with very few images of atypical postures or extreme viewpoints \cite{huang2019followmeup}.
This data bias makes models trained with this dataset vulnerable to not detecting postures not included in the training data (e.g., occlusion or extreme viewpoints) \cite{jain2019robustness, li2020yoloacn}.
Such shape-dependency constraints can cause performance degradation in real-world environments where the target shape is not guaranteed, such as in life-saving operations at disaster sites \cite{valarmathi2023disaster_yolo, papyan2024ai_drone_rescue, nihal2024uav} or surveillance tasks where the target's posture is unpredictable.
While research is underway to address these challenges within the CNN framework through data augmentation and model architecture improvements, this paper describes a complementary approach based on physical principles.

This study aimed to bridge this gap through system-level optimization to translate physical principles into an efficient system that operates under real-world constraints.
The design decisions to narrow down the discriminatively effective spectral information to four bands and to adopt a lightweight multi-layer perceptron (MLP) eliminate the need for specialized sensors or GPUs, enabling real-time edge implementation.
This paper makes the following contributions:

\begin{enumerate}
    \item Implementation and Evaluation on Edge Hardware: The 
4-band multispectral method \cite{ono2025jimaging} was implemented on a physical camera system and 
its performance without a GPU edge device was evaluated.
The implementation 
achieved 23.2 frames per second (fps), providing evidence for its real-time utility 
under resource constraints.
\item Validation of Complementary Value: The author quantitatively evaluated the effectiveness of the clothing-based method in extreme posture changes and simulated disaster rescue scenarios where CNN-based detectors show performance degradation, and assessed its value to complement existing sensing technologies.
\item Achievement of Practical Performance: The practicality of the clothing-based approach was demonstrated by the construction of a prototype using only commercially available components and a single-board computer (Raspberry Pi 5; Raspberry Pi Ltd., Cambridge, UK), achieving a real-time performance of 23.2 fps.
\item Development of an Evaluation Methodology: The author analyzed the detection characteristics of the clothing-based method using physical principles and demonstrated the validity of the evaluation metrics tailored to the task definition of "presence detection" based on data.
\end{enumerate}

\section{Related Work}

\subsection{Spectral Analysis for Material Identification}
The method described here is based on the principle of identifying material-specific spectral reflectance characteristics (spectral signatures), which is a well-established technique in the field of remote sensing \cite{clark1999spectroscopy_review}.
The effectiveness of this principle has been demonstrated in applications such as mineral resource exploration through satellite observation \cite{goetz1985imaging_spectrometry}.
Furthermore, one of the technologies applying this principle is hyperspectral anomaly detection (HAD).
HAD is a technology that detects objects with different spectral characteristics from the background without prior information and is expected to be applied to search and rescue (SAR) and surveillance activities.
However, according to several survey papers, conventional HAD research has shown challenges in high-dimensional data processing involving hundreds of bands, such as high computational cost, sensitivity to noise, and limited generalization performance \cite{su2022had_survey, shah2022hyperspectral, zhao2015global}.
Due to these characteristics, conventional HAD systems require expensive and complex sensors and large computational resources, making them unsuitable for edge applications that demand real-time performance and low power consumption.

In contrast, the principle of spectral analysis is also applied in more familiar industrial fields.
In particular, near-infrared spectroscopy has attracted attention as a technology for non-destructive identification of fabric composition in the textile industry \cite{ciurczak2021nir_handbook}, and the characteristic spectra of common fibers such as cotton and polyester are known to be classifiable \cite{paz2024discrimination}.
Thus, while the effectiveness of the identification principle based on spectral information itself has been demonstrated in various fields, it is essential to bridge the gap between "principle" and "practicality," such as that faced by HAD, to apply it to actual edge computing applications.

\subsection{Multispectral Fusion for Person Detection}
In the field of computer vision, multispectral methods that fuse visible (VIS) and thermal infrared (thermal) images are being researched to achieve robust person detection.
These methods use the complementarity of information from both modalities: VIS images provide detailed texture information during the day under good lighting conditions, while thermal images, which capture the thermal radiation of objects, facilitate separation from the background at night or in adverse weather \cite{hwang2015multispectral, li2019illumination}.
Various fusion architectures have been proposed to leverage this complementarity, achieving success particularly in improving nighttime detection performance \cite{nie2024multispectral}.
However, many of these approaches ultimately input the enriched features from the fusion into a CNN-based architecture to perform detection by recognizing spatial patterns (i.e., shape) \cite{li2019illumination}.
Therefore, even if the dimensionality of the input information increases, the recognition principle inherently shares the shape-dependency challenge common to CNN-based methods.

\section{Description of HitoMi-Cam}
\subsection{Principle: Spectral Signatures of Clothing}

\begin{figure}[H]
\centering
\includegraphics[width=1.0\textwidth]{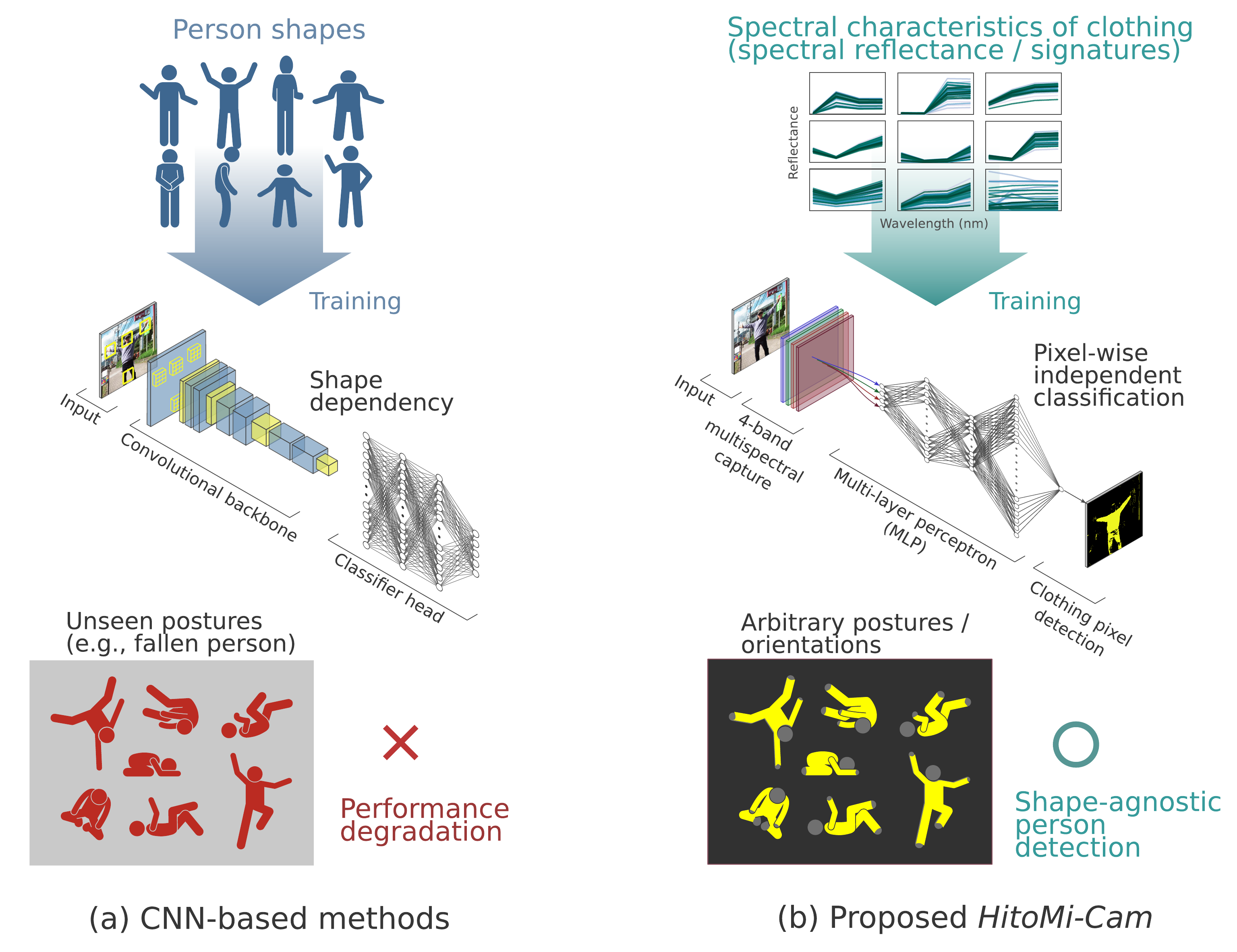}
\caption{Conceptual comparison of detection principles of \hitomi{} and CNN-based methods.
(a) Conventional methods such as Convolutional Neural Networks (CNNs) depend on the "shape" patterns of people included in the training data.
Therefore, they face the challenge of shape dependency, where performance degrades for postures not present in the training data (e.g., a fallen person).
(b) \hitomi{} focuses on the physical spectral reflectance characteristics (spectral signatures) of clothing materials rather than the shape of the object.
By classifying each pixel independently, it aims for shape-agnostic detection that does not depend on the target's posture or orientation.}
\label{fig:concept}
\end{figure}

To address the aforementioned challenges to practical implementation, this study builds upon our previous simulation study published in this journal~\cite{ono2025jimaging}, which identified four spectral bands effective for clothing identification.
This work focuses on translating that theoretical foundation into a practical system.
The physical basis of this method is the prior knowledge that common clothing fibers (polyester, cotton, and wool) exhibit characteristic reflectance in specific near-infrared wavelength regions under outdoor daylight conditions~\cite{zhao2019spectral, cura2021textile_recognition}.
These fibers are reported to account for about 80\% of the global textile market share \cite{textile_exchange2022report}.
This property allows the discrimination of clothing versus non-clothing at the pixel level to function as a proxy for determining the presence of a person.
The author's previous research identified four effective wavelength bands for discrimination (central wavelengths of 457, 565, 645, and 735 nm) \cite{ono2025jimaging}.
This approach differs from existing CNN-based methods, which rely on geometric information such as shape, orientation, and posture, as it exclusively uses material-specific spectral information (see Figure~\ref{fig:concept}).

\subsection{System Architecture}
\label{sec:system_arch}
\begin{figure}[H]
\centering
\includegraphics[width=1.0\textwidth]{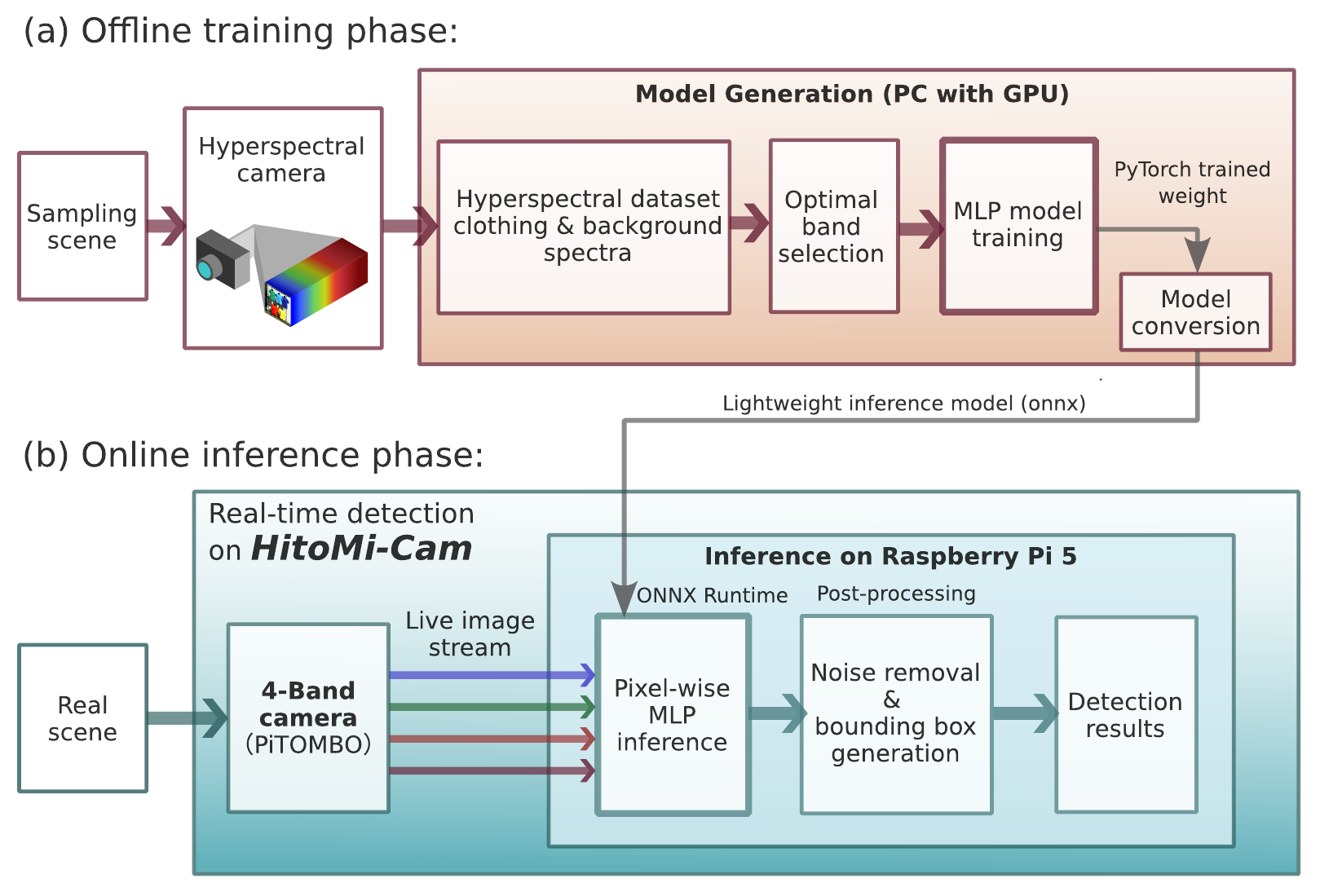}
\caption{Overall system architecture of \hitomi{}. The system consists of two tiers: (a) an offline learning phase and (b) an online inference phase.
In (a), based on the methods and datasets established in the author's previous research \cite{ono2025jimaging}, a 4-band selection is made from hyperspectral data on a GPU-equipped PC, a multi-layer perceptron (MLP) model is trained, and an Open Neural Network Exchange (ONNX) format inference model is generated.
In (b), the \hitomi{} prototype, equipped with the generated inference model, performs person detection on a Raspberry Pi for real-time input from a 4-band camera.}
\label{fig:architecture}
\end{figure}

To realize the concept shown in Figure~\ref{fig:concept}, the author designed a two-tiered architecture consisting of an offline learning system and an online inference system (see Figure~\ref{fig:architecture}).

The offline learning system (Figure~\ref{fig:architecture}a) runs on a PC equipped with a GPU.
Here, a lightweight MLP model is trained using the PyTorch framework based on a spectral dataset (84 types of clothing and 35 types of background) collected with a hyperspectral camera.
The data are converted to the Open Neural Network Exchange (ONNX) format after the model is trained to generate an inference model suitable for deployment on edge devices.
The online inference system (Figure~\ref{fig:architecture}b) is \hitomi{}. This system loads the pre-trained ONNX model generated offline onto a Raspberry Pi 5 and performs inference on each pixel using the ONNX Runtime engine on the Raspberry Pi to detect the presence of a person.
This two-tiered architecture reflects the design philosophy of completing the computationally intensive model training offline, enabling high-speed, lightweight inference on resource-constrained edge devices.
The adoption of a 4-band multispectral method, which forms the core of this architecture, and a lightweight MLP as the classifier is based on the analysis in previous research \cite{ono2025jimaging}.
In that study, evaluation using 167-band hyperspectral data showed that this 4-band configuration is appropriate in terms of the balance between computational efficiency and detection performance, and that an MLP demonstrates sufficient performance for its classification.
In this study, considering further computational efficiency on edge devices, the hidden layers were optimized to a 2-layer configuration with 16 and 8 nodes.
\subsection{\hitomi{} Camera System Prototype}
\label{sec:prototype}
\begin{figure}[H]
\centering
\includegraphics[width=0.6\textwidth]{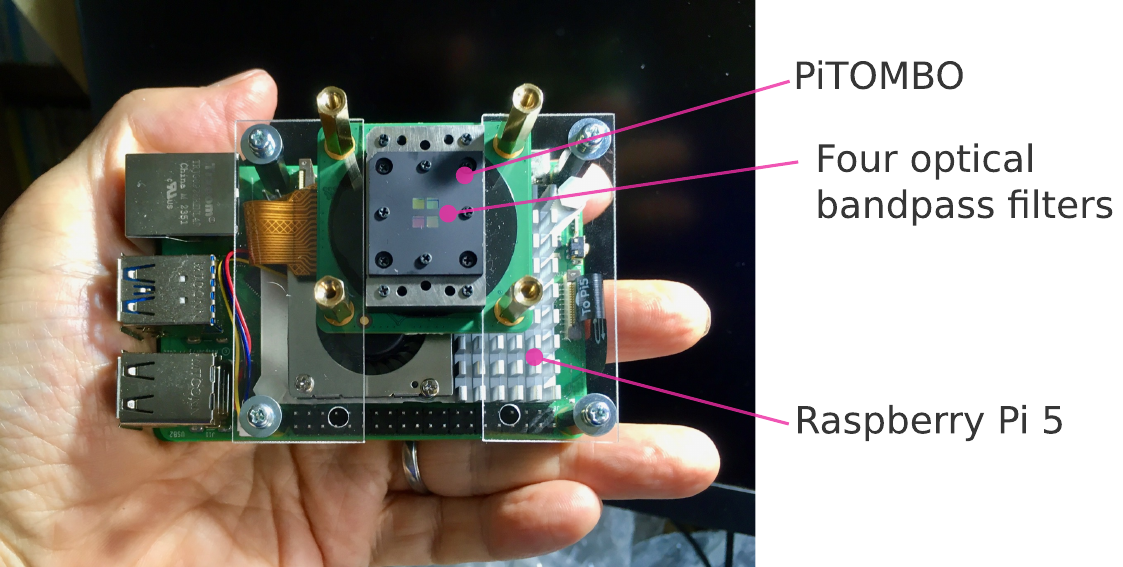}
\caption{Physical configuration of the \hitomi{} system. The system consists of a single-board computer (Raspberry Pi 5) as the host computer and a commercially available compound-eye camera module (PiTOMBO; Asahi Electronics Laboratory Co., Ltd., Osaka, Japan).
As shown, the camera module is equipped with four optical bandpass filters and can simultaneously acquire a multispectral image with four spectral bands (central wavelengths 457, 565, 645, and 735 nm) in a single shot.}
\label{fig:system}
\end{figure}

To demonstrate the clothing-based method, the author built a camera system named \hitomi{} (see Figure~\ref{fig:system}).
The system is composed of a commercially available compound-eye camera module (PiTOMBO; Asahi Electronics Laboratory Co.,  Ltd., Osaka, Japan) equipped with four lenses and optical bandpass filters \cite{nakanishi2020tombo} and a single-board computer (Raspberry Pi 5) serving as the host.

The selection of this camera module was guided by the practical requirements of the prototype. It was chosen as it offered a good balance of the combined criteria: a compact form, simultaneous 4-band imaging, Raspberry Pi compatibility, and control via Python libraries.
The selection of the Raspberry Pi 5 as the edge device was based on recent benchmark studies showing that it can achieve real-time performance for lightweight computer vision tasks \cite{minott2025edge_benchmark, alqahtani2024benchmarking}.

A primary objective of this study was to verify whether \hitomi{} could achieve real-time performance on an edge device. Preliminary experiments investigated the trade-off between processing speed and image resolution by comparing various sensor operation modes (full-size vs. binning) and downsampling ratios. Based on these experiments, the author selected a resolution of 253 $\times$ 190 pixels, captured in binning mode, as this configuration approached real-time processing speeds (approx. 23.2 fps) without significantly compromising detection performance. This decision was based on the hypothesis that the pixel-wise classification nature of the clothing-based method would be relatively robust to reductions in resolution. Detailed processing time data supporting this choice are provided in the Supplementary Material (Table S-1).

The selection of the four optical filters for this prototype was based on constraints for practical application. While the theoretically optimal wavelength set identified in prior research \cite{ono2025jimaging} assumed narrow bands, for a physical camera implementation, ensuring sufficient light intensity through adequate bandwidth and the commercial availability of filters are critical factors. That same study also confirmed that the chosen wavelength set is robust against shifts in the central wavelength and increases in bandwidth.
The four bands installed in this unit have central wavelengths of 457, 565, 645, and 735 nm, with corresponding full width at half maximum values of 36, 25, 21, and 29 nm, respectively. This band set was determined by reapplying the search method established in the author's prior work, with the added constraints of commercial filter availability and light intensity. 

For strict reproducibility, other factors are also important, including the transmission characteristics of the filters used, the spectral sensitivity (quantum efficiency curve) of the image sensor, exposure and gain settings during capture, and the specific white balance (WB) correction procedure described in Section 3.3. However, this paper reports the primary parameters of central wavelengths and bandwidths.
This re-searched band set was adopted after confirming in simulations that its performance remained close to the theoretical solution. The validity of this choice is also empirically supported by the 84.5\% recall rate obtained with the physical device in the basic verification experiment (Section 3.1).

This configuration enables the acquisition of a 4-band multispectral image in a single shot. By avoiding the need for a dedicated sensor or GPU, it allows for a compact, low-cost implementation, which is advantageous for applications such as drone-based deployment.
It should be noted that the compound-eye camera used in this system is subject to parallax due to the physical distance between its lenses. Furthermore, misalignments among the four images can occur from errors in the optical axis alignment of the lenses. These effects are suppressed to an average inter-band misalignment of less than one pixel by pre-calibrating the pixel shift amounts between bands and applying a correction process. However, because the image sensor in this prototype uses a rolling shutter, a slight time lag occurs in the image acquisition timing between the four bands. The effect is suppressed to an acceptable level by operating the sensor in a high-frame-rate binning mode. Under the experimental conditions of this study, this effect was considered negligible, although it could become a potential issue when imaging faster-moving subjects.

\subsection{Clothing Map Generation Algorithm}
\begin{figure}[H]
\centering
\includegraphics[width=0.6\textwidth]{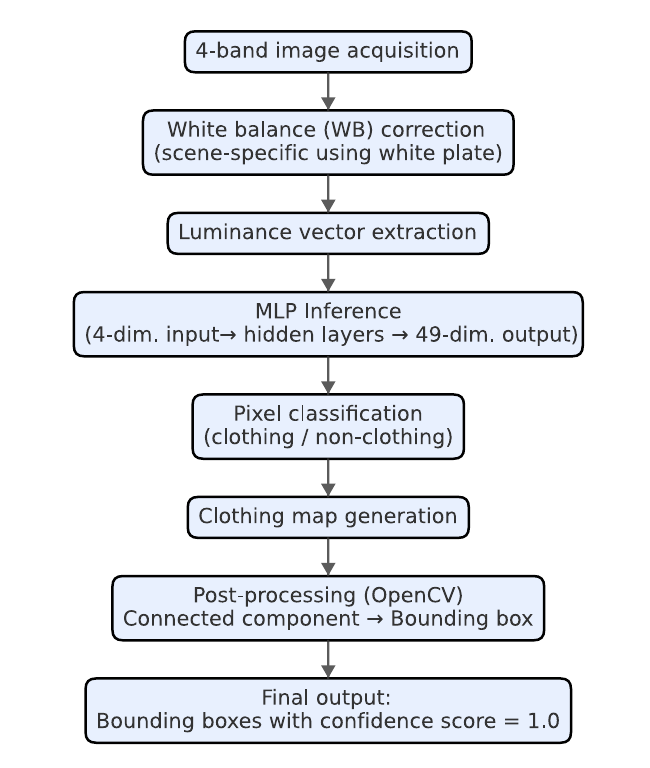}
\caption{Processing pipeline from 4-band multispectral image acquisition to clothing map generation in \hitomi{}.
The process starts with the acquisition of a 4-band multispectral image.
After the luminance vector of each pixel is extracted, it is input into a lightweight MLP.
The MLP independently classifies each pixel as "clothing" or "non-clothing" to generate an initial clothing map.
Subsequently, post-processing using OpenCV (noise removal and morphological operations) is applied to finalize the clothing regions and calculate the final bounding boxes.}
\label{fig:pipeline}
\end{figure}

The \hitomi{} detection algorithm is designed to minimize computational load (see Figure~\ref{fig:pipeline}).

The processing flow begins by applying a WB correction to the acquired 4-band image to compensate for variations in the lighting environment. The stability of this method, which is based on spectral information, depends on the spectral composition of the illumination, making this correction crucial. Specifically, a standard white plate is photographed immediately before shooting each evaluation scene, and correction coefficients are calculated and recorded so that the luminance data of the white plate area are flat across the four bands. Applying these scene-specific correction coefficients ensures that the input data are consistent with the spectral characteristics assumed by the trained model, thereby guaranteeing the robustness of the subsequent MLP inference.

A luminance vector is extracted from the corrected image and input into the pre-trained MLP. The classifier selection and training process is based on the methods established in previous research \cite{ono2025jimaging}. In that study, several machine learning models were comparatively evaluated, and it was concluded that the MLP is suitable for this task due to its balance between classification performance and computational efficiency.
The MLP used in this study was built using the PyTorch Lightning framework. The model takes a 4-dimensional spectral vector as input, passes it through two hidden layers (16 and 8 nodes with the ReLU activation function), and outputs a 49-dimensional classification vector. 

The training dataset was generated by performing hyperspectral measurements on 84 common types of clothing and 35 background materials, and then simulating the 4-band spectral signatures used in this study from that data. 
A detailed list of the clothing samples used for training and the distribution of their spectral characteristics are provided in the Supplementary Material (Table S-2 and Figure S-1), based on the author's prior work.
 
The Adam optimizer (learning rate of 0.001) was used for training, with a setting to stop the training early if the validation loss (monitored on a validation set, which was randomly sampled from the spectral dataset) did not improve for 2 epochs (maximum number of epochs: 500).

In addition, data augmentation similar to that in the previous study was performed to improve the diversity and robustness of the training data. Color chart images were added to the background data to reinforce the samples of artificial objects, and the luminance was varied by a factor ranging from 0.68 to 1.46 to enhance the resilience against lighting fluctuations. During training, sampling was adjusted to balance the total number of pixels in the "clothing" and "non-clothing (non-organic and plant)" categories.

The trained MLP performs binary classification on each pixel independently, assigning it a label of "clothing" or "non-clothing." More specifically, for each input pixel's spectral vector, the MLP outputs a 49-dimensional score vector. This output corresponds to the 39 clothing categories and 2 background categories (a total of 41 classes) defined in the previous research. During inference, the class with the highest score in this vector is adopted, and based on whether that class belongs to a clothing category, each pixel is finally classified as "clothing" or "non-clothing." Note that the MLP was initially designed for 49 labels but was ultimately consolidated to 41 labels. 	As a result, 8 units are not effectively used. However, because the impact on the learning results was minor, this configuration was maintained for the model in this study.

\begin{figure}[H]
\centering
\includegraphics[width=1.0\textwidth]{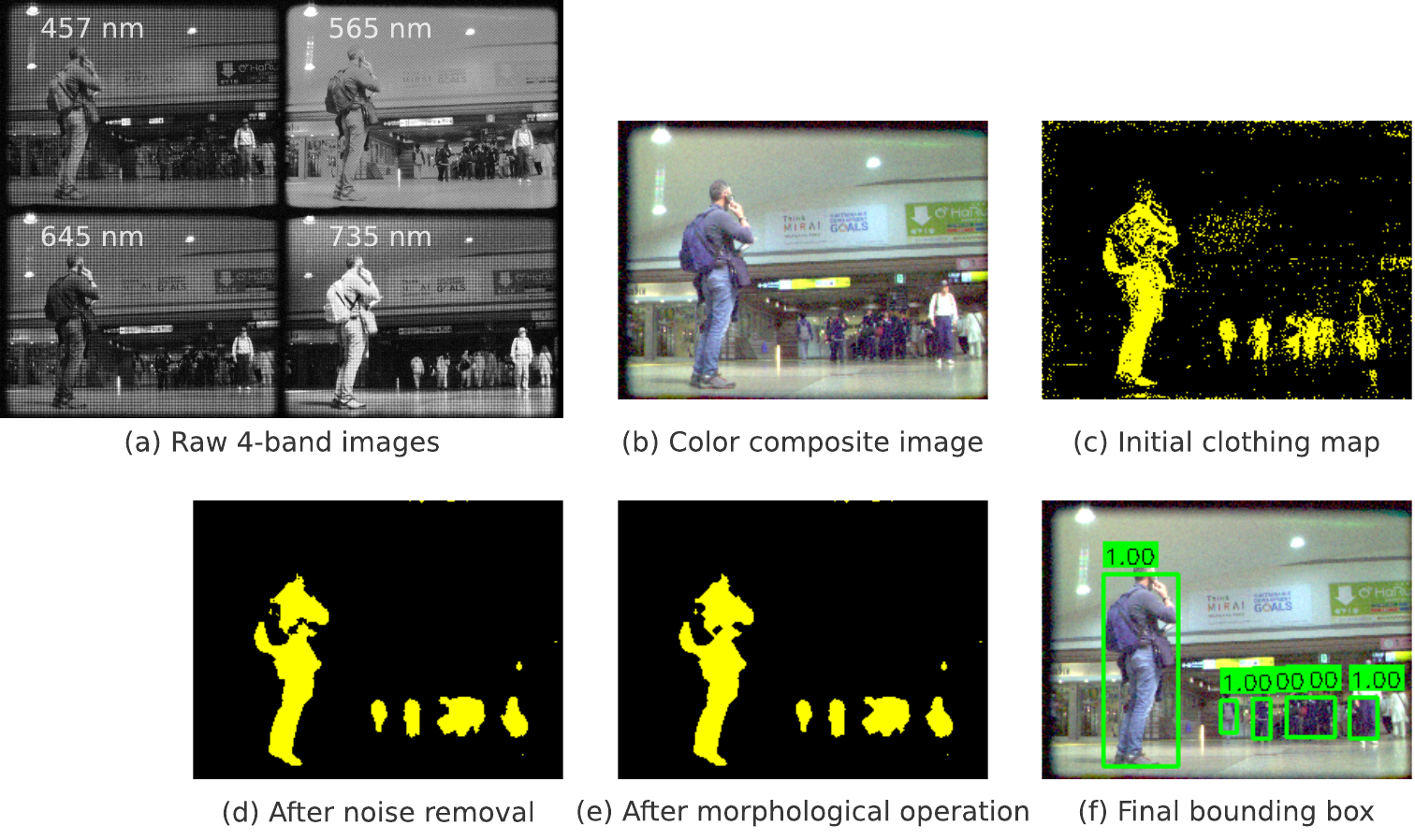}
\caption{Example of entire processing sequence from 4-band image acquisition to final bounding box generation.
(a) Raw 4-band images captured by camera. (b) Color composite image synthesized for easy viewing.
The band assignments are R=645, G=565, and B=457 nm. (c) Initial map of pixels classified as "clothing" by MLP.
(d) Map after noise removal. (e) Map after morphological operations are applied to define continuous regions.
(f) Final bounding box calculated for defined clothing region.}
\label{fig:processing_example}
\end{figure}

This pixel-wise processing design reduces the computational cost and, in principle, aims for detection that is independent of the target's shape or rotation. For each connected component in the resulting "clothing map," a bounding box is fitted using OpenCV in a post-processing step (see Figure~\ref{fig:processing_example}). The output of \hitomi{} consists of the positional information of this bounding box and a confidence score of 1.0, based on the physical detection. This characteristic differs from many CNN-based detectors, which output a continuous confidence score ranging from a value greater than 0.0 to 1.0 as a statistical probability.

\section{Evaluation}
\subsection{Basic Verification: Physical Reproducibility with Principle}
\label{sec:basic_verify}
To confirm that the \hitomi{} prototype could physically implement the clothing-based principle whose effectiveness was demonstrated based on simulations in previous research \cite{ono2025jimaging}, the author conducted a basic verification.
The 84 clothing type samples used in the previous study were photographed with the prototype under clear outdoor conditions, and 71 of them were detected as clothing (recall rate of 84.5\%).
This result indicates that the detection principle based on spectral information shown in the simulation was physically reproduced by the constructed hardware (see Supplementary Material Figure S-2).
\subsection{Evaluation Policy: Defining a Presence Detection Task}
\label{sec:eval_policy}
To evaluate the performance of \hitomi{}, it was necessary to define the target task not as precise object localization but rather as presence detection. The goal is to answer the question "Is there a possibility that a person is present in this location?" in applications like disaster rescue with high reliability while minimizing the waste of resources due to false detections.

\begin{figure}[H]
\centering
\includegraphics[width=1.0\textwidth]{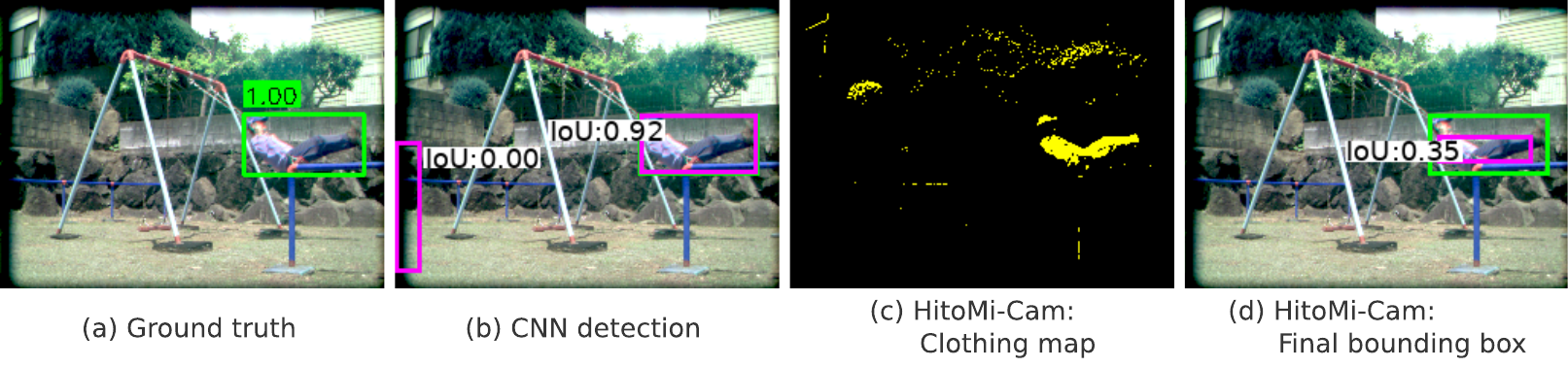}
\caption{Bounding box generation characteristics of \hitomi{} and validation of evaluation metric's appropriateness. (a) Ground truth region derived from manual annotation.
(b) A conventional CNN-based detector (YOLOv5x) yielded a result with high spatial agreement with the ground truth (Intersection over Union (IoU) = 0.92).
(c) In contrast, \hitomi{}'s principle is to first generate a map of pixels classified as "clothing."
(d) A bounding box is subsequently fitted to this clothing map, resulting in a detection that is systematically smaller than the ground truth (IoU = 0.35 in this example).
This process illustrates that, while the detection may be considered a failure at a standard threshold (e.g., IoU $\geq$ 0.5), it is successful for the task of presence detection.
Therefore, this analysis supports the use of a lower threshold, such as IoU = 0.2, for appropriately evaluating methods with distinct detection principles.}
\label{fig:iou_justification}
\end{figure}

This task setting justifies the use of a low intersection over union (IoU) threshold. \hitomi{}'s detection principle, which involves first generating a map of pixels classified as "clothing" and subsequently fitting a bounding box to this map, inherently leads to bounding boxes that are systematically smaller than the ground truth regions encompassing the entire person (as illustrated in Figure~\ref{fig:iou_justification}). Consequently, a threshold that does not demand excessive spatial accuracy is necessary to properly evaluate this characteristic. This evaluation approach is practically viable because the system generates few false positives (FP), as discussed in Section 3.5.4, where \hitomi{} shows a FP rate lower than that of CNNs. Therefore, the choice of IoU = 0.2 as the primary evaluation threshold is a principle-based selection for assessing the task of presence detection, a strategy built upon this inherent low FP characteristic.

\subsection{Experimental Setup}
\label{sec:setup}

The COCO pre-trained models (YOLOv5, etc.) are used as established baselines for shape-based detection. The goal of this comparison is not to claim superiority over these models, but rather to identify the specific scenarios (i.e., Simulated SAR, Swing) where their performance degrades, and to demonstrate the complementary effectiveness of the proposed approach in those exact contexts.

To verify the effectiveness of the clothing-based method, the author compared the following three series of models: (1) \hitomi{}, (2) the YOLOv5 family (n, s, m, l, and x) as representatives of high-precision CNN-based detectors, and (3) the relatively classic MobileNet-V1 and EfficientDet-L0 as relatively lightweight CNNs focused on speed.

In the evaluation of all models, the confidence threshold was unified to 0.01. This low threshold was chosen primarily to enable a fair comparison of the recall rates across models, as it ensures that even low-confidence candidates from the CNN-based models are considered. Furthermore, this setting reflects the realistic condition required in operation for detectors to achieve a high recall rate in applications like SAR, where "misses (false negatives [FN]) cannot be tolerated."   Note that, because \hitomi{} outputs only candidates with a confidence of 1.0, this threshold setting effectively means all its detections are included.

For the compared CNN models, the following common policy was used to ensure fairness. The YOLOv5, EfficientDet-L0, and MobileNet-V1 models all used the pre-trained weights officially provided for the MS COCO dataset. No additional training or fine-tuning was performed on the evaluation dataset. For these models, the output was filtered to include only the "person" class (COCO class index 0). The implementation for inference considered the standard execution environment of each model. The YOLOv5 family was implemented directly on the CPU using the PyTorch framework, and EfficientDet-L0 and MobileNet-V1 used the TensorFlow Lite runtime optimized for execution on edge devices. For the evaluation of detection performance (average precision [AP] calculation), the color images (253 $\times$ 190 pixels) generated in the image shaping process of \hitomi{} were used as input to ensure that all models were evaluated on images with the same content. Each CNN model automatically resized this low-resolution image to the model-specific required resolution (e.g., 640 $\times$ 640 for YOLOv5s) in its internal pre-processing for inference. No special optimizations, such as FP16 quantization, were applied.

\subsection{Performance Evaluation Dataset}
\label{sec:dataset}
To evaluate the detection performance of \hitomi{} in a real-world environment, the author constructed three types of datasets (see Table~\ref{tab:datasets}).
These are image sequences of actual people, independent of the spectral measurement data used for training.
The ground truth bounding boxes were manually annotated by the author.
\begin{table}[H]
\caption{Overview of evaluation datasets.}
\label{tab:datasets}
\begin{adjustwidth}{-\extralength}{0cm}
\begin{tabularx}{\fulllength}{X c c c X}
\toprule
\textbf{\begin{tabular}{@{}c@{}}Dataset \\ name\end{tabular}} & 
\textbf{\begin{tabular}{@{}c@{}}Frame \\ count\end{tabular}} & 
\textbf{Environment} & 
\textbf{\begin{tabular}{@{}c@{}}Shooting \\ period\end{tabular}} & 
\textbf{\begin{tabular}{@{}c@{}}Main \\ features\end{tabular}} \\
\midrule
General Scene & 1200 & Outdoor/Indoor & May 2025 & Diverse person scenes \\
Simulated SAR Scene & 160 & Outdoor & August 2025 & Complex postures, occlusion \\
Swing Scene & 300 & Outdoor & May 2025 & High-speed motion, extreme posture changes \\
\bottomrule
\end{tabularx}
\end{adjustwidth}
\end{table}

Each dataset was designed with a different evaluation purpose. The General Scene dataset targets people in typical standing postures and is used to evaluate baseline performance. The Simulated SAR Scene and Swing Scene datasets assume static and dynamic situations where the shape becomes atypical, respectively, and are intended to evaluate robustness under conditions where CNN performance may deteriorate.

To clarify the dataset design and generalization testing strategy: (1) The General Scene dataset was captured in public spaces with multiple, non-specific subjects under diverse lighting conditions, including outdoor (direct sunlight and shade) and indoor (mixed artificial light and skylight) environments, establishing baseline performance across typical postures. (2) The Swing Scene features a single subject (the author) dynamically changing posture, testing robustness to extreme shape variations. (3) The Simulated SAR Scene was designed to simulate realistic search-and-rescue conditions by placing clothing samples on the ground with atypical shapes in complex backgrounds including vegetation with high near-infrared reflectance similar to that of clothing and blue tarps commonly found at disaster sites. Critically, both the Swing Scene and Simulated SAR Scene use clothing samples not included in the training dataset, enabling evaluation of generalizability to unseen materials. Additionally, the Simulated SAR Scene incorporates multiple spectral challenges: some clothing was wetted with water to alter spectral characteristics, lighting conditions were varied by including both sunny and shady areas within frames, and multi-angle overhead perspectives were captured to simulate drone-based search operations. This design creates a challenging yet realistic scenario that provides atypical (rather than zero) shape cues, enabling fair evaluation of the proposed method's complementarity under conditions where typical CNN shape-recognition struggles.

\begin{figure}[H]
\centering
\includegraphics[width=1.0\textwidth]{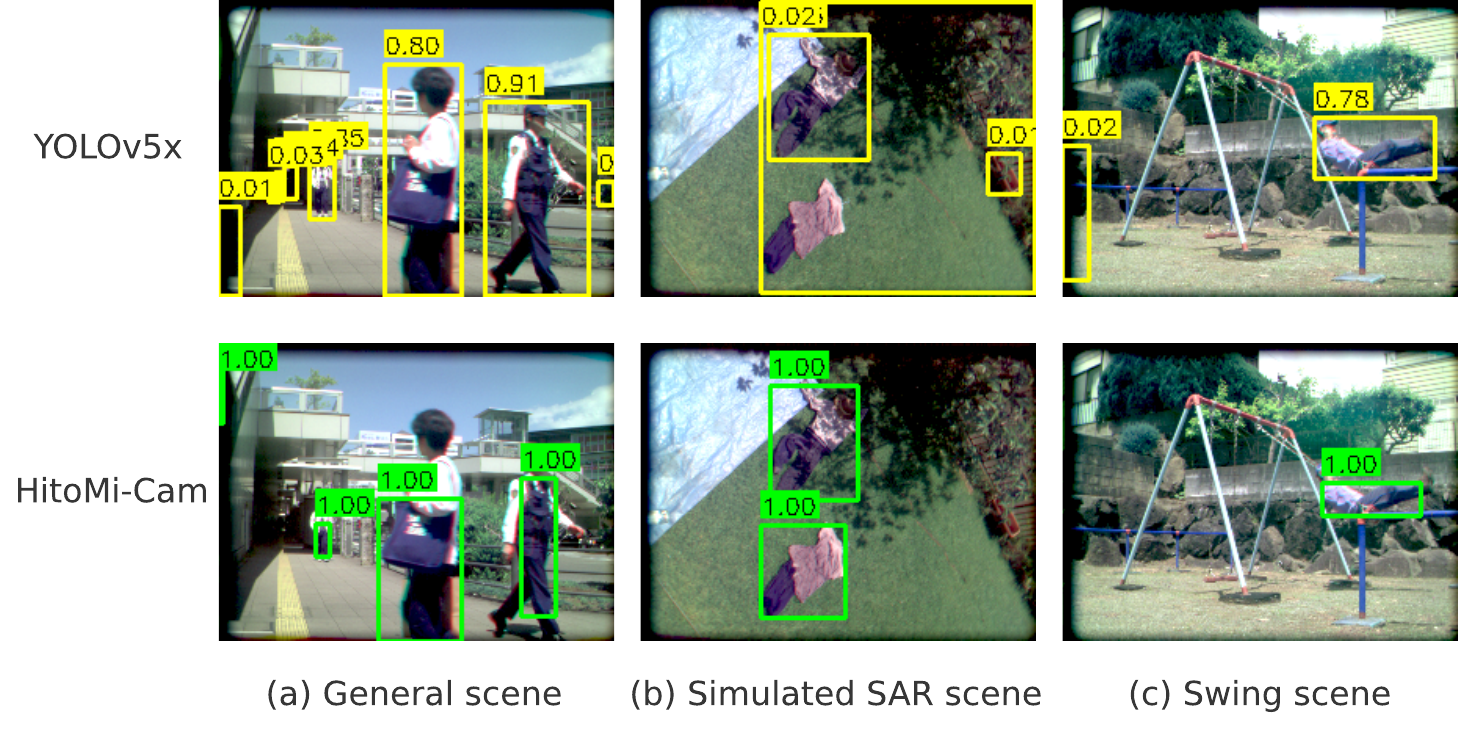}
\caption{Comparison of representative frames and detection results from the three datasets.
Yellow indicates results from the comparison method (YOLOv5x), and green indicates \hitomi{} results.
(a) In the case of General Scene, the CNN (yellow) shows high accuracy, while \hitomi{} (green) detects only the clothing area, generating a smaller rectangle.
(b) In the case of Simulated SAR Scene, detection by the CNN (yellow) fails because the target does not have a human shape, but \hitomi{} (green) captures the material characteristics and detects the ground truth.
(c) In the Swing Scene, the CNN's (yellow) detection becomes unstable for atypical postures, while the shape-agnostic \hitomi{} (green) continues to track the target.}
\label{fig:dataset_examples}
\end{figure}

An overview of each dataset is given in Table~\ref{tab:datasets}. Representative frames from each dataset and examples of detection results by HitoMi-Cam and the comparison methods are shown in Figure~\ref{fig:dataset_examples}. 

For a showcase of qualitative results for each scenario, see Figures S-3–S-5 in the Supplementary Material. Notably, as shown in the Supplementary Material (e.g., Figure S-4), the system demonstrates robustness even in frames with significant motion blur, a condition expected in practical SAR deployment.

\subsection{Results}
\label{sec:results}

The performance metrics reported for \hitomi{} in this section (e.g., Table~\ref{tab:results}) are based on a single, optimized model. As discussed in Section~\ref{sec:interpretation}, this model was intentionally selected from multiple training attempts (42 runs) to best align with the method's design philosophy of minimizing false positives. A full statistical analysis of performance variance (e.g., standard deviation) from training randomness is a limitation of this study and is noted as an important area for future work.

The quantitative comparison results for the three scenarios are summarized in Table~\ref{tab:results}, with detailed metrics for all models provided in Supplementary Material Table S-3.
The details of each scenario are described below.

\begin{table}[H]
\caption{Average precision (AP) of each model across three evaluation scenarios, evaluated at an Intersection over Union (IoU) threshold of 0.2.}
\label{tab:results}
\centering
\begin{tabular}{cccc}
\toprule
\textbf{Model} & \textbf{General Scene} & \textbf{Simulated SAR Scene} & \textbf{Swing Scene} \\
\midrule
\hitomi{} & 0.340 & 0.935 & 0.957 \\
EfficientDet-L0 & 0.780 & 0.370 & 0.733 \\
MobileNet-V1 & 0.688 & 0.409 & 0.459 \\
YOLOv5n & 0.864 & 0.506 & 0.762 \\
YOLOv5s & 0.936 & 0.520 & 0.930 \\
YOLOv5m & 0.961 & 0.524 & 0.974 \\
YOLOv5l & 0.968 & 0.538 & 0.970 \\
YOLOv5x & 0.978 & 0.536 & 0.976 \\
\bottomrule
\end{tabular}
\end{table}

\subsubsection{Scenario 1: Baseline Performance in General 
Scenes}
As a baseline, the performance on General Scene images was evaluated.
The performance at AP@IoU = 0.2 reached 0.86 for YOLOv5n and 0.98 for YOLOv5x, while \hitomi{}'s was 0.34, confirming that CNN models are effective in environments with people having typical shapes.
This result also suggests the limitations of \hitomi{}. That is, because the method detects people by capturing the difference in spectral characteristics from the background, it cannot detect clothing with spectral characteristics very similar to the background.
Furthermore, two other phenomena frequently occurred in this scenario, leading to decreased performance.
First, due to subtle differences in spectral characteristics between upper and lower body clothing, often only one part was detected, resulting in systematically smaller bounding boxes.
Second, when multiple people were close together, the algorithm tended to merge them into a single connected component, failing to count them as multiple true positives (TP).
These specific failure cases are visually demonstrated in Figure S-6 of the Supplementary Material.
\subsubsection{Scenario 2: Effectiveness in Simulated SAR Scenes}
\begin{figure}[H]
\centering
\includegraphics[width=0.7\textwidth]{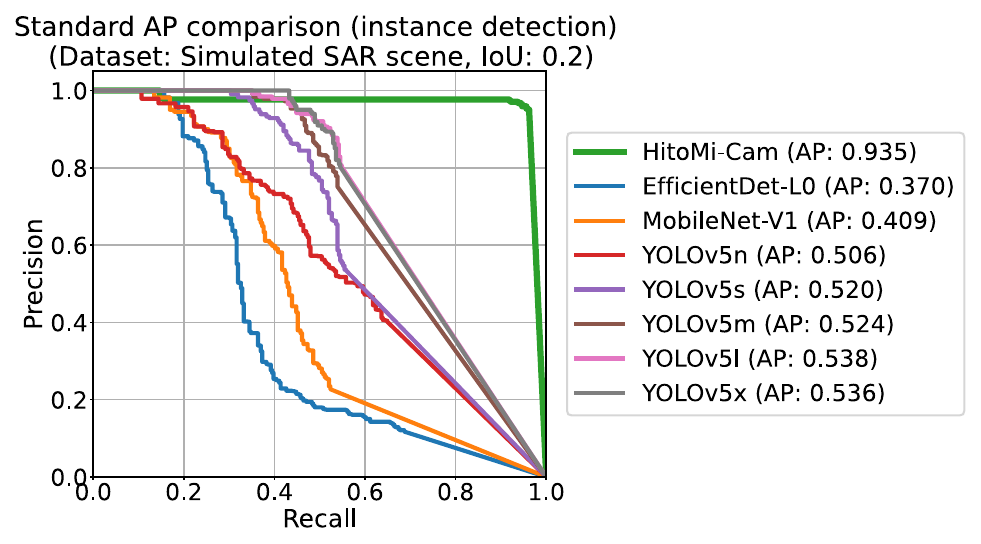}
\caption{Precision-recall (PR) curves for each model in the Simulated SAR Scene (IoU $\geq$ 0.2).
The horizontal axis represents recall, and the vertical axis represents precision; curves closer to the top-right indicate better performance.
Because \hitomi{} outputs only candidates with a confidence score of 1.0 based on physical detection, its PR curve is rendered as a horizontal line.
This contrasts with the CNN-based models, which form curves corresponding to varying confidence thresholds.
The Average Precision (AP) for \hitomi{} was 93.5\%, while the best-performing CNN model (YOLOv5l) achieved 53.8\%.}
\label{fig:pr_curve_sar}
\end{figure}

The effectiveness of \hitomi{} was confirmed in scenarios where no typical human shape is found.
The evaluation result in the Simulated SAR Scene images showed that \hitomi{}'s AP was 93.5\%, while the best comparison model (YOLOv5l) had an AP of 53.8\%.
Figure~\ref{fig:pr_curve_sar} shows the precision-recall (PR) curves of each model in this scenario.
\hitomi{}'s curve (blue line), located in the upper right of the graph, showed superior performance compared to the other CNN models.
(For the PR curves of the other two scenes, see Supplementary Material Figure S-9.)

In contrast, the performance of the YOLOv5 family was understandably limited in this scenario.
This is not a failure of the models themselves but rather an expected outcome of the task definition, which required classifying atypical, non-human-like shapes as the "person" class.
Such targets represent a significant deviation from the typical postures learned from the MS COCO dataset, making the task exceptionally challenging for any shape-based detector.
A detailed analysis of this behavior, presented in the Supplementary Material (Figure S-8), reveals that the models often misclassified the targets as other objects with similar abstract shapes (e.g., "dog" or "kite").
This result highlights the fundamental difference in principle and showcases a scenario where \hitomi{}'s shape-agnostic approach, based on "clothing pixel identification," can serve as a valuable complement to conventional methods that rely on "human shape."
\subsubsection{Scenario 3: Robustness to Extreme Posture Changes (Swing Scenes)}
\begin{figure}[H]
\centering
\includegraphics[width=0.9\textwidth]{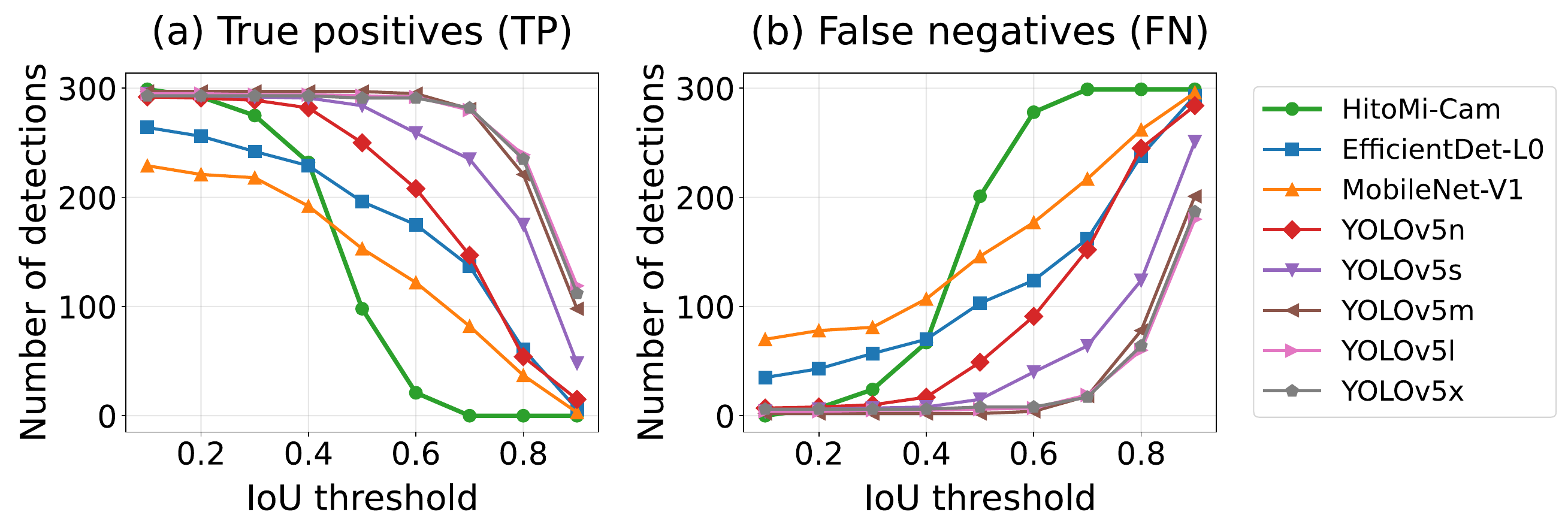}
\caption{Transition of (a) True Positives (TP) and (b) False Negatives (FN) with respect to IoU threshold changes for Swing Scene images.
The horizontal axis represents the IoU threshold, and the vertical axis represents the number of detections.
\hitomi{} maintains a high TP rate and low FN rate up to IoU = 0.3, but above 0.4, the TP rate sharply declines while the FN rate increases dramatically.
This behavior is consistent with the method's characteristic of generating smaller detection boxes, suggesting that evaluation at high IoU thresholds may be inappropriate for this method and supporting the choice of IoU = 0.2 as the threshold.}
\label{fig:tp_fn_swing}
\end{figure}

\hitomi{} demonstrated stable detection performance even in Swing Scene images, where a person's shape changes dynamically.
Analyzing the performance transition of each model with respect to the IoU threshold (Figure~\ref{fig:tp_fn_swing}, and Supplementary Material Table S-4 and Figure S-10) showed that \hitomi{}'s behavior differs for each scene.
In the Swing and General Scene images, ground truth annotations encompass entire human figures, whereas \hitomi{} detects only clothing regions, resulting in systematically smaller detection boxes.
Consequently, IoU values concentrate around 0.3–0.4, with the TP rate sharply declining when the threshold exceeds 0.3.
Simultaneously, the FN rate remains low up to IoU = 0.3 but increases dramatically beyond 0.4.
This behavior is consistent with the method’s characteristic tendency to generate systematically smaller detection boxes.
In contrast, in the simulated SAR scenes, the ground truth annotation was set only to the clothing area, so the detection target and detection principle matched.
This allowed \hitomi{} to maintain high IoU values and stable performance even at IoU = 0.5.
This analysis suggests that low IoU thresholds are appropriate for evaluating realistic scenarios (swing and general scenes) and that, under certain conditions (simulated SAR scenes), \hitomi{} has the potential to achieve high positional accuracy.

\subsubsection{False Positive Characteristics}
\label{sec:fp_char}
Under the evaluation condition with a confidence threshold set to 0.01, \hitomi{} consistently exhibited a lower number of FP than did the CNN models across all three scenario datasets.
This can be attributed to the differing operational principles: \hitomi{} outputs only candidates with a confidence of 1.0 based on physical spectral information, whereas CNNs generate numerous candidates, including those with low confidence scores, based on statistical probabilities (see Supplementary Material Figure S-11).
\subsubsection{Processing Performance}
Table~\ref{tab:processing_time_breakdown} shows a detailed breakdown of the average processing time of \hitomi{} on the Raspberry Pi 5 (mean).
Table~\ref{tab:end_to_end_time} shows the end-to-end processing time of each model on the Raspberry Pi 5. In measuring the processing time of each CNN model, standard JPEG images were acquired directly from the camera module.
The end-to-end average processing time of \hitomi{} was 43.1 ms (23.2 fps).
This time includes pre-processing such as data shaping from raw images (16.6 ms) and post-processing (9.5 ms).
The core process of the method, that is, inference by the MLP, is completed in 14.0 ms.
This indicates that the proposed algorithm is computationally efficient, with bottlenecks in the pre- and post-processing stages.
Conversely, in the case of CNN models, there is almost no overhead for pre- and post-processing.
Nevertheless, many CNN models require more processing time. Among the CNN series, only MobileNet-V1 achieved a speed close to real time (65.1 ms).
The YOLOv5 series required 213 ms for the n model and 2720 ms for the x model.
\begin{table}[H]
\centering
\caption{Detailed breakdown of \hitomi{}'s end-to-end average processing time (Raspberry Pi 5).
Of the total processing time, the inference stage accounts for only 31\%.
The majority of the current processing time is occupied by pre-processing (41\%) and post-processing (25\%).}
\label{tab:processing_time_breakdown}
\begin{tabular}{llccc}
\toprule
\textbf{Stage} & \textbf{Sub-Stage} & \textbf{\begin{tabular}{@{}c@{}}Time \\ (ms)\end{tabular}} & \textbf{\begin{tabular}{@{}c@{}}Subtotal \\ (ms)\end{tabular}} & \textbf{\begin{tabular}{@{}c@{}}Total \\ (ms)\end{tabular}} \\
\midrule
Capture & & \multicolumn{2}{c}{0.03} & \multirow{10}{*}{\textbf{43.1}} \\
\cmidrule(r){1-4}
\multirow{3}{*}{Pre-processing} & 4-band Image Acquisition & 12.0 & 17.8 & \\
 & WB Correction & 4.5 & & \\
 & Luminance Vector Extraction & 1.2 & & \\
\cmidrule(r){1-4}
\multirow{2}{*}{Inference} & MLP Inference & 9.8 & 13.4 & \\
 & Pixel Classification & 3.7 & & \\
\cmidrule(r){1-4}
\multirow{3}{*}{Post-processing} & Clothing Map Generation & 0.8 & 10.8 & \\
 & Post-processing (OpenCV) & 5.5 & & \\
 & Final 
Output & 4.4 & & \\
\cmidrule(r){1-4}
Others & & \multicolumn{2}{c}{1.1} & \\
\bottomrule
\end{tabular}
\end{table}

\begin{table}[H]
\caption{End-to-end average processing time of each model (Raspberry Pi 5).
This table shows the end-to-end processing time (ms) measured on the Raspberry Pi 5 for each model and the corresponding frame rate (fps).}
\label{tab:end_to_end_time}
\centering
\begin{tabular}{ccc}
\toprule
\textbf{Model} & \textbf{Time (ms)} & \textbf{fps} \\
\midrule
\hitomi{} & 43.1 & 23.2 \\
MobileNet-V1 & 65.1 & 15.4 \\
EfficientDet-L0 & 131.0 & 7.6 \\
YOLOv5n & 212.7 & 4.7 \\
YOLOv5s & 420.0 & 2.4 \\
YOLOv5m & 871.3 & 1.1 \\
YOLOv5l & 1667.6 & 0.6 \\
YOLOv5x & 2719.9 & 0.4 \\
\bottomrule
\end{tabular}
\end{table}

\begin{figure}[H]
\centering
\includegraphics[width=1.0\textwidth]{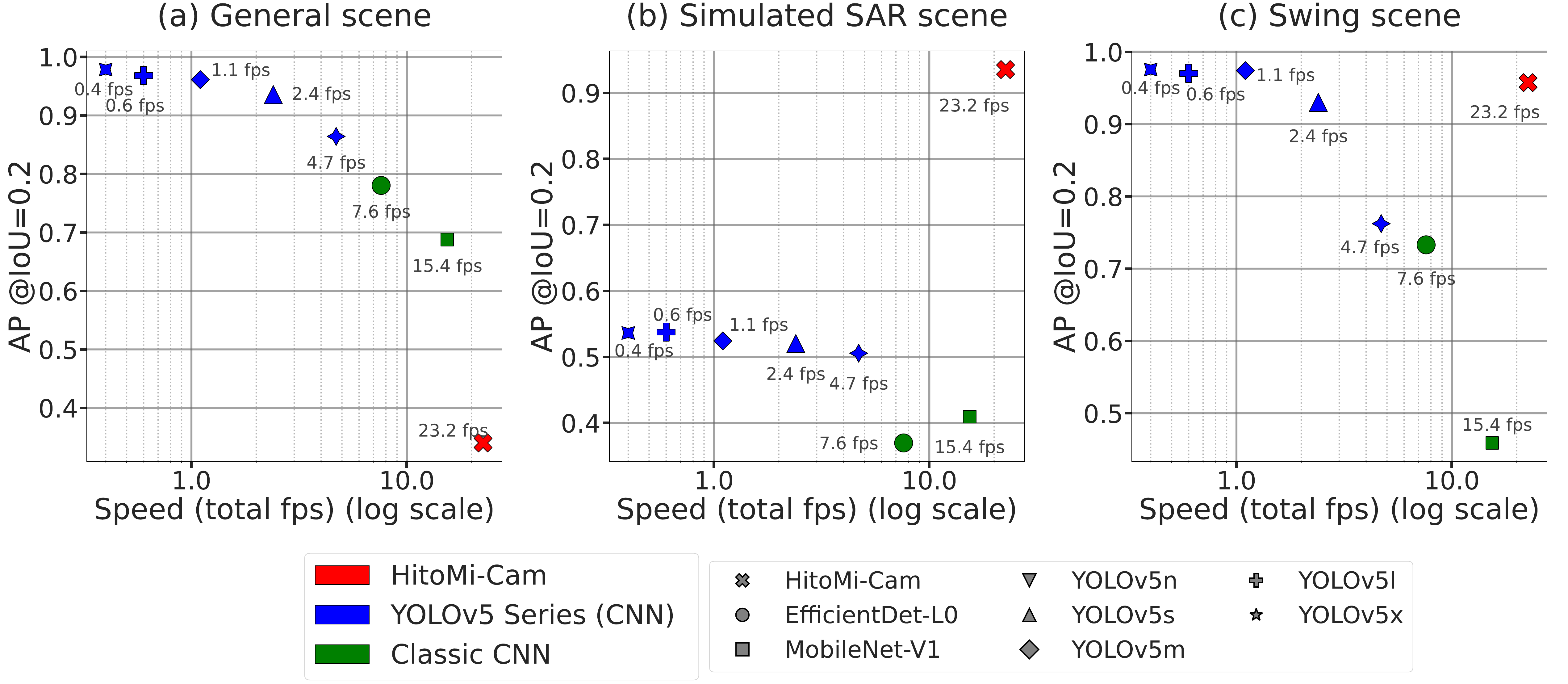}
\caption{Trade-off relationship between performance (AP@IoU $\geq$ 0.2) and processing speed (fps) of each method in three scenarios.
(a) For the General Scene, the CNN-based models occupy the high-performance, low-speed region (upper left), while \hitomi{} is located in the low-performance, high-speed region (lower right).
For the (b) Simulated SAR Scene and (c) the Swing Scene, \hitomi{} is located in the high-performance, high-speed region (upper right).
These results show that under specific conditions where the performance of CNNs deteriorates, \hitomi{} achieves both processing speed and detection accuracy.}
\label{fig:performance_speed_tradeoff}
\end{figure}

Figure~\ref{fig:performance_speed_tradeoff} shows the relationship between the performance (AP@IoU$\geq$0.2) and inference speed (fps) of each method for the three scenario datasets.
From this figure, the characteristics of the three series, \hitomi{} (red), YOLOv5 series (blue), and lightweight CNN series (green), can be clearly distinguished.
The inference speed of \hitomi{} is located in a clearly faster region compared to all CNN models.

Furthermore, to confirm the actual operation on an edge device, the supplementary material video (Video S-1) shows a parallel comparison of four models: \hitomi{}, YOLOv5n, MobileNet-V1, and EfficientDet-L0.
This video visually confirms that \hitomi{} can maintain high-speed and stable detection even in difficult dynamic scenes.
\section{Discussion}
\subsection{Interpretation of Results and Advantages of P Method}
\label{sec:interpretation}
The experimental results show that \hitomi{} and CNN-based methods have a trade-off relationship based on their different recognition principles and function in a mutually complementary manner.
While CNNs showed superiority for general scenes, \hitomi{} demonstrated its effectiveness for scenarios where the typical human shape is compromised, such as in simulated SAR scenes and swing scenes.
Table~\ref{tab:conceptual_comparison} shows a conceptual comparison.

\begin{table}[H]
\caption{Conceptual comparison of conventional CNN-based detectors and \hitomi{}.}
\label{tab:conceptual_comparison}
\begin{adjustwidth}{-\extralength}{0cm}
\begin{tabularx}{\fulllength}{l X X}
\toprule
\textbf{Feature} & \textbf{Conventional CNN-based Detectors (e.g., YOLO)} & \textbf{\hitomi{}} \\
\midrule
\textbf{Basic Principle} & Statistical pattern recognition & Physics-based material classification \\
\textbf{Main Information Source} & Spatial patterns of intensity/color (shape and texture) & Spectral reflectance characteristics (material composition) \\
\textbf{Dependency} & Depends on the diversity and bias of the training dataset & Depends on the spectral characteristics of the material, but not on the posture or shape \\
\textbf{Core Task Definition} & Shape recognition: "Where is the person's outline?"
& Presence detection: "Is there a person?" \\
\textbf{Main Failure Factor} & Atypical postures/shapes not in the training data & Materials with spectral characteristics similar to those of the background \\
\textbf{Output Characteristics} & Generates numerous candidates with positive, continuous confidence scores up to 1.0 & Generates a few candidates, each with a confidence score of 1.0 \\
\textbf{Optimal Environment} & Structured environments where typical human shapes can be predicted & Unstructured environments where shape cues are unreliable \\
\textbf{Nature of False Positives} & Statistical error (trade-off with confidence threshold) & Physical misclassification (similarity of spectral characteristics of materials) \\
\bottomrule
\end{tabularx}
\end{adjustwidth}
\end{table}

One of the characteristics arising 
from this difference in principle is the behavior of FP generation.
CNN-based detectors generally control the balance between precision and recall by adjusting the confidence threshold during inference.
If the threshold is lowered to prioritize recall, detection misses decrease, but a trade-off arises where misdetections based on the similarity of statistical patterns increase.

In contrast, the low FP rate of \hitomi{} shown in this paper reflects not only the difference in recognition principle but also the intentional design philosophy of the clothing-based approach.
The author considered variations due to initial learning values and selected a model from multiple training attempts that minimizes background misdetection while maximizing the identification performance of clothing pixels.
In other words, rather than adjusting the trade-off during inference time, as CNNs do, the classifier itself is designed with an inherent "high-precision (low-misdetection)" characteristic.
The advantages of this method are also clarified by comparison with VIS+thermal fusion systems.
These methods are effective in that they make the "shape" of a person stand out from the background by capturing thermal radiation, but the recognition principle depends on shape information.
In contrast, \hitomi{} depends on the spectral characteristics of materials, an information source different from shape and temperature, and the author believes it shows a new direction for sensing that complements existing multimodal approaches.

Another key advantage of this principle is its inherent robustness to motion blur. CNN-based methods rely heavily on spatial shape cues (e.g., edges), which are severely degraded by motion blur. In contrast, this method relies on the pixel-wise spectral signature (material), which is more resilient to such artifacts as the underlying material properties remain detectable. The stable performance in blurred frames, noted in Section~\ref{sec:dataset} and shown in the Supplementary Material (Figure S-4), is considered a direct result of this principled advantage.

\subsection{Limitations}
The physical principle of this method is the source of its effectiveness, but it also defines its limitations.
\subsubsection{Dependence on Clothing as aProxy}
Because this method is based on the premise of "using the presence of clothing as a proxy for the presence of a person," its effectiveness may be diminished in situations with minimal clothing coverage.
\subsubsection{Spectral Ambiguity}
Because the detection principle depends on the difference in spectral characteristics between the target and the background, performance limitations arise when it is difficult to distinguish between the two.
\subsubsection{Dependence on Ambient Light and Contaminants}
Because this method is a passive sensing system that uses ambient light, such as sunlight, for illumination, its performance depends on the available ambient light.
Its effectiveness at night or in indoor environments has not been verified.
Furthermore, the presence of environmental contaminants such as mud, water, and dust in disaster rescue scenarios changes the spectral reflectance characteristics of clothing and becomes a factor that degrades detection performance.
\subsubsection{Potential for Misdetection Due to Non-clothing Objects}
Although not apparent in the experimental environment of this study, the possibility of misdetecting non-clothing objects with spectral characteristics similar to clothing exists in principle.

For example, in an environment like a clothing store or shopping mall, the method would, by design, generate numerous false positives from static, non-person clothing.

\subsubsection{Image Sensor Constraints}
The image sensor used in this prototype is a color sensor with an RGB Bayer filter on each pixel.
From the perspective of sensitivity and resolution, a monochrome sensor is ideal, and this configuration is the result of prioritizing the availability of commercial modules.
\subsection{Practical Value and Application Fields}
Considering the technical characteristics and limitations of this method, its practical application fields include scenarios where the target shape is unreliable, such as disaster rescue and search for missing persons.
In these situations, the typical human shape that is a premise for CNNs may not be evident.
However, the shape-agnostic detection principle of \hitomi{} can discover targets that CNNs might miss by capturing the physical cues of clothing. 
This directly contributes to preventing misses (reducing False Negatives, FN), which is the most critical requirement in SAR operations.

At the same time, the low False Positive (FP) rate demonstrated in this study provides significant practical value.
In SAR activities, a high FP rate, while less critical than a single FN, leads to wasted resources. It forces rescue teams to investigate numerous false alarms, slowing down the overall search and ultimately reducing the total area that can be covered, thereby lowering the overall probability of finding a victim. \hitomi{}'s dual advantages---its ability to complement the FNs of shape-based detectors and its low-FP characteristic for maintaining operational efficiency---can function as highly reliable information to assist an operator's judgment.

Furthermore, the prototype developed in this study itself has practical value. Traditionally, collecting spectral data required professional-grade, bulky, and expensive equipment (e.g., hyperspectral imagers), making large-scale, in-field dataset construction difficult. A key contribution of this work is the demonstration of \hitomi{} as a compact, low-cost, and energy-efficient edge device. This prototype effectively lowers the barrier to data collection, positioning it as an essential 'first step' that enables the extensive future validation called for by this field.

\section{Conclusion and Future Work}
Building on our previous work published in this journal~\cite{ono2025jimaging}, which established the theoretical foundation through simulation, this study implemented and evaluated a practical camera system on physical hardware.
The experimental verification demonstrated that the spectral-based approach can operate effectively under real-time constraints and resource limitations of edge computing environments.
The experimental verification in this study revealed that, while the detection performance of this method is not as good as that of CNNs in many situations, it has the characteristics of achieving a low FP rate and high-speed processing on edge devices in detecting humans based on pixel-level clothing identification.
This "high-speed and low FP rate" characteristic creates practical value in complementing existing shape-dependent detectors in disaster SAR for missing persons.

This study bridges the gap between theoretical principles and practical deployment.
While prior work \cite{ono2025jimaging} established the theoretical foundation through simulation, the present findings demonstrate that the spectral-based principle can operate under the real-time constraints and resource limitations of edge computing environments.

Future work is expected to cover the following points:

\begin{enumerate}
\item Fusion with CNN-based spatial pattern recognition: A hierarchical approach can be considered where \hitomi{} quickly generates candidates for "clothing-like" regions, and CNNs are applied only to those regions.

\item Performance improvement in nighttime and varied environments: Further quantitative analysis under various passive lighting conditions (e.g., indoor tungsten, cloudy skies) is also needed. To overcome these dependencies entirely, integrating an active illumination system that selectively irradiates specific near-infrared wavelength bands, operation regardless of day or night is expected to be possible. Overcoming the limitations of passive sensing in this manner is a high-priority future task to expand the method's practical application fields.

\item Hardware improvements and validation: Further reduction in power consumption, miniaturization, introduction of a global shutter type image sensor (to eliminate rolling shutter artifacts), and signal-to-noise ratio improvement by employing a monochrome sensor can be considered. This also includes a more detailed quantitative reporting of calibration error metrics (e.g., pixel shift residuals) beyond the average misalignment reported in Section~\ref{sec:prototype}.

\item Systematic evaluation and countermeasures for misdetection due to non-clothing objects: Misdetection due to non-clothing objects with spectral characteristics similar to those of clothing may become a problem when \hitomi{} is deployed in various real environments.

To address this issue, it is necessary to catalog background objects with clothing-like spectra and construct an evaluation dataset based on them. As a more practical countermeasure for static objects (such as in a clothing store), fusing this method with simple motion analysis to filter out non-moving targets would be an effective strategy.

\item Improving localization accuracy: While the current method focuses on "presence detection" (as discussed in Section~\ref{sec:eval_policy}), its systematically smaller bounding boxes are a limitation for tasks requiring more accurate localization. It is acknowledged that how to reliably estimate a full-body region from only the clothing-pixel map is a non-trivial challenge that warrants dedicated future study.

\item Statistical robustness analysis: As noted in Section~\ref{sec:results}, the current performance is reported from a single, selected model. A full statistical analysis of model performance variance (e.g., standard deviation) arising from training randomness, using methods such as cross-validation or repeated training runs, is an important task for future validation.

Comprehensive benchmarking against state-of-the-art (SOTA) detectors: As noted in Section~\ref{sec:setup}, the current comparison was designed to demonstrate complementarity with established baselines rather than to comprehensively cover all detector families. The field of real-time object detection continues to evolve, with recent architectures such as YOLOv8~\cite{terven2023yolo_review}—featuring an anchor-free detection head and architectural refinements—and RT-DETR (Real-Time DEtection TRansformer)~\cite{zhao2024detrs}—a transformer-based detector achieving real-time performance without non-maximum suppression—representing current advances in SOTA detection. A broader benchmark incorporating these and other contemporary detectors is acknowledged as an important future task to more comprehensively contextualize this method's performance profile and validate its complementary role across diverse detection paradigms.

\item Model architecture optimization: The current MLP configuration (16 and 8 nodes) was chosen based on the findings in prior work \cite{ono2025jimaging} and further optimized for computational efficiency on the edge device (as noted in Section~\ref{sec:system_arch}). A systematic ablation study on the trade-off between model complexity and detection accuracy was not performed and is acknowledged as a subject for future study.

\end{enumerate}

These developments are expected to contribute to the construction of person detection systems.

\section*{Author Contributions}
Conceptualization, S.O.; methodology, S.O.; software, S.O.; validation, S.O.; formal analysis, S.O.; investigation, S.O.; resources, S.O.; data curation, S.O.; writing---original draft preparation, S.O.; writing---review and editing, S.O.; visualization, S.O.; supervision, S.O.; project administration, S.O.; funding acquisition, S.O. The author has read and agreed to the published version of the manuscript.

\section*{Funding}
This research received no external funding.

\section*{Institutional Review Board Statement}
Not applicable.

\section*{Informed Consent Statement}
Informed consent was obtained from the primary subject who participated in the controlled data acquisition scenarios (e.g., the swing and clothing sample photography). For datasets recorded in public spaces (e.g., the general scene at the station), individuals were captured incidentally. The collected footage has been processed in such a way that individuals are not personally identifiable.

\section*{Data Availability Statement}
The data presented in this study are available on request from the corresponding author. The data are not publicly available due to privacy restrictions.

\section*{Acknowledgments}
The author thanks the FUJIFILM Corporation for providing the research environment and technical support.

\section*{Conflicts of Interest}
The author declares no conflicts of interest.

\section*{Abbreviations}
The following abbreviations are used in this manuscript:

\begin{tabular}{@{}ll}
AP & Average precision\\
CNN & Convolutional neural network\\
FN & False Negative\\
FP & False Positive\\
HAD & Hyperspectral anomaly detection\\
IoU & Intersection over union\\
MLP & Multi-layer perceptron\\
ONNX & Open Neural Network Exchange\\ 
PR & Precision-recall\\
SAR & Search and rescue\\
TP & True Positive\\
VIS & Visible\\
WB & White balance\\
YOLO & You Only Look Once\\ 
\end{tabular}

\clearpage       
\appendix

\setcounter{figure}{0}
\setcounter{table}{0}
\renewcommand{\thefigure}{S-\arabic{figure}}
\renewcommand{\thetable}{S-\arabic{table}}

\captionsetup[tabularx]{singlelinecheck=false, justification=raggedright}

\begin{center}
  {\LARGE \textbf{Supplementary Material for \\ HitoMi-Cam: A Shape-Agnostic Person Detection Method Using the Spectral Characteristics of Clothing}} \\[1em]
  {\large Shuji Ono}
\end{center}
\vspace{1cm}

\section*{Supplementary Figures}

\begin{figure}[H]
  \centering
  \includegraphics[width=0.95\textwidth]{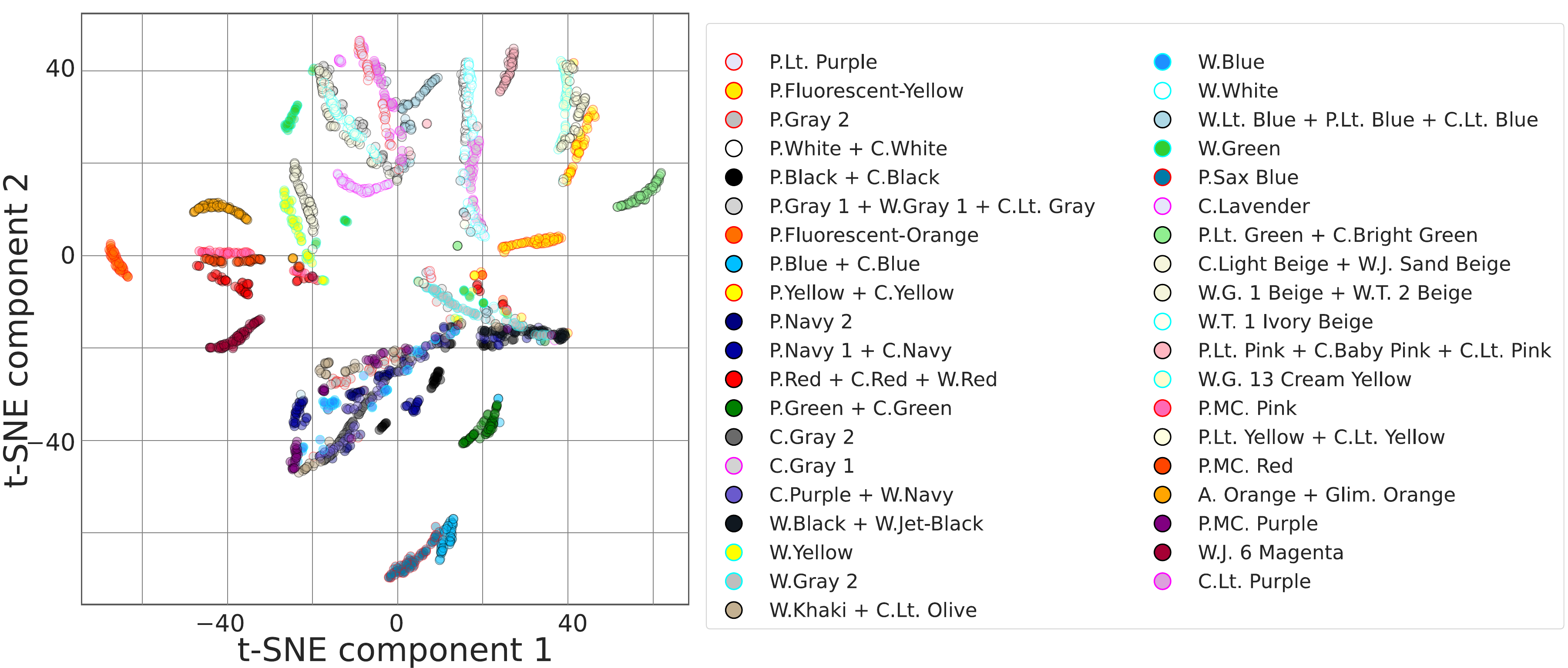}
  \caption{{t-Distributed stochastic neighbor embedding (t-SNE) visualization of hyperspectral signals for 39 labeled clothing categories. Each point represents a clothing sample, plotted in a two-dimensional space based on its 167-band hyperspectral signal. The distribution illustrates the spectral diversity of the clothing samples used in the training dataset. This figure is adapted from the author's previous work.}}
  \label{fig2-3-3}

  \vspace{2mm} 
  \parbox{0.95\textwidth}{
    \footnotesize{
    \textsuperscript{1} Key to abbreviations: P. = polyester, 
    C. = cotton,
    W. = wool,
    T. = Toray (a Japanese synthetic fabric manufacturer), G. = gabardine,
    MC. = mixed color,
    J. = Josette (a linen-based fabric), A. = athletic,
    Glim. = glimmer,
    and Lt. = light.
    }
  }
\end{figure}

\begin{figure}[H]
\centering
\includegraphics[width=1.0\textwidth]{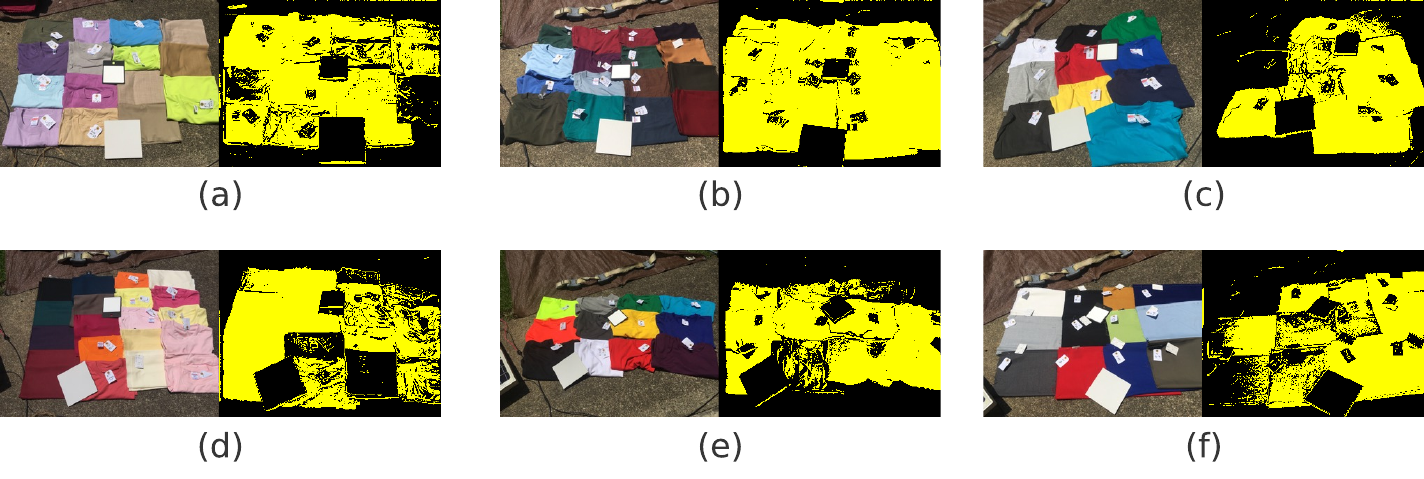}
\caption{Example of results of basic verification experiment with \hitomi{} prototype. Representative examples of the 84 types of clothing samples whose spectral characteristics were measured in the previous study were photographed under natural outdoor light to verify the detection performance. (a)--(g) show the detection results for each sample, with the yellow area indicating the pixel map correctly detected as "clothing" by \hitomi{}. Of the 84 types, 71 were correctly detected (a recall rate of 84.5\%), demonstrating that the detection principle based on spectral information shown in the simulation is physically reproduced by the constructed hardware.}
\label{fig:supp_basic_verification}
\end{figure}

\begin{figure}[H]
\centering
\includegraphics[width=1.0\textwidth]{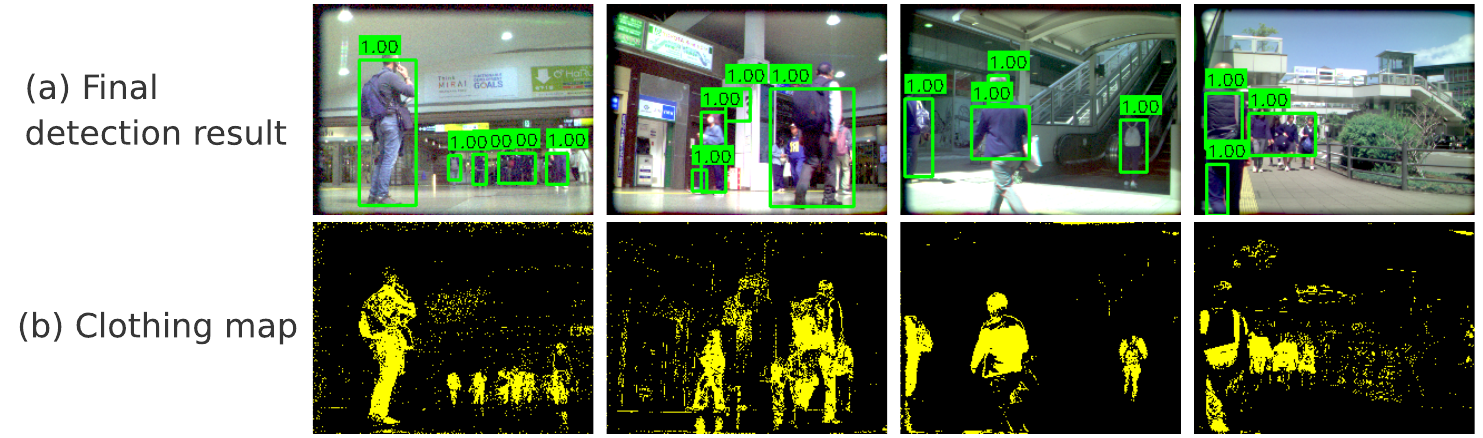}
\caption{Qualitative results of \hitomi{} in the General Scene. (a) Examples of final detection results. (b) The corresponding examples of intermediate clothing maps.}\label{fig:supp_basic_verification_2g} 
\end{figure}

\begin{figure}[H]
\centering
\includegraphics[width=1.0\textwidth]{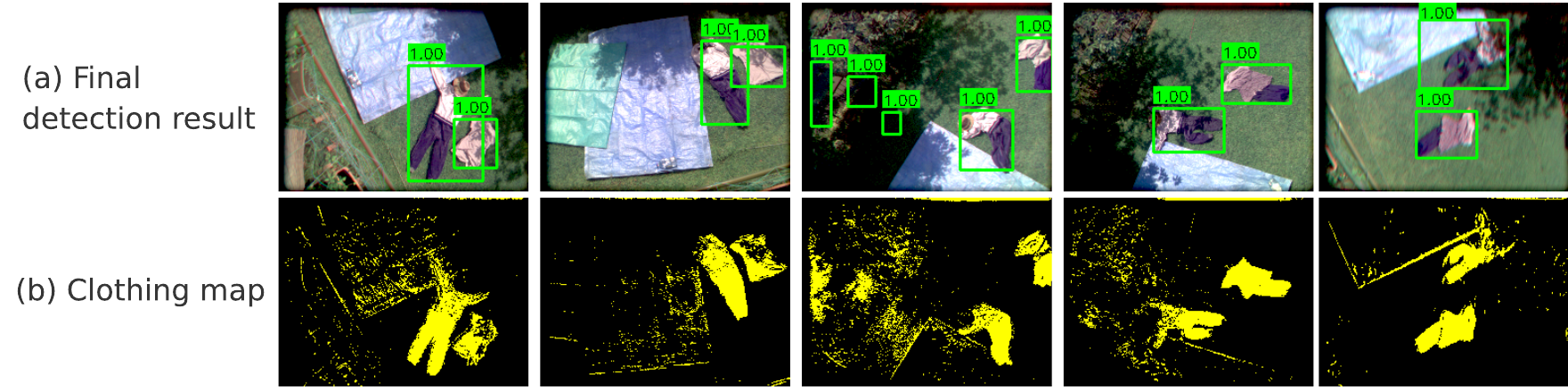}
\caption{Qualitative results of \hitomi{} in the Simulated SAR Scene. (a) Examples of final detection results. (b) The corresponding sequence of intermediate clothing maps. The examples include various challenges, such as complex backgrounds (vegetation, blue tarp) and lighting (shade). Notably, the rightmost example demonstrates robustness to significant motion blur, a condition that typically degrades the performance of shape-based detectors.}\label{fig:supp_basic_verification_SAR} 
\end{figure}

\begin{figure}[H]
\centering
\includegraphics[width=1.0\textwidth]{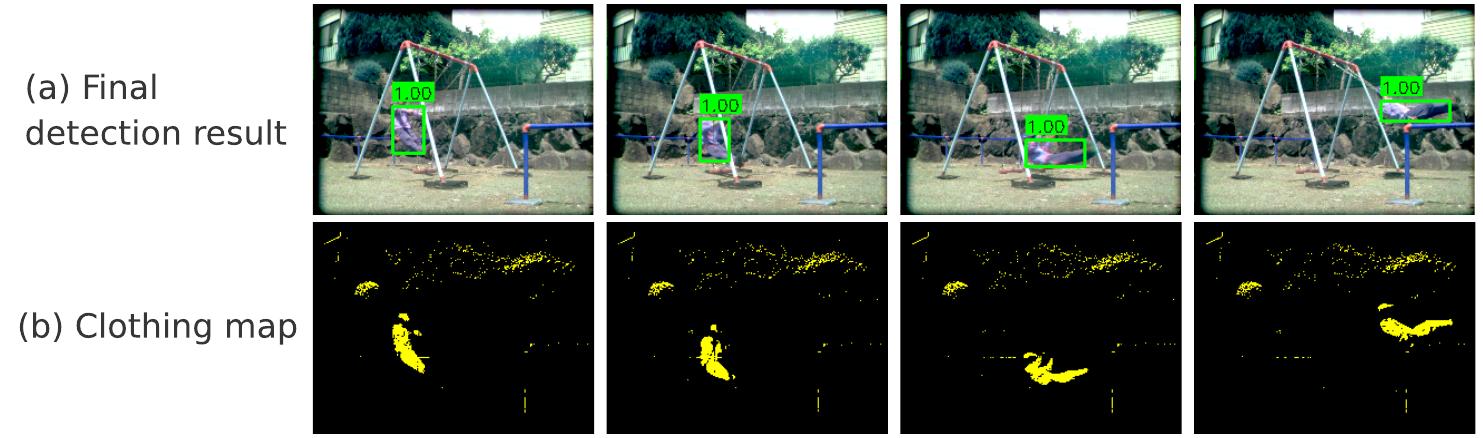}
\caption{Qualitative results of \hitomi{} in the Swing Scene. (a) A sequence of final detection results. (b) The corresponding sequence of intermediate clothing maps.}\label{fig:supp_basic_verification_2s} 
\end{figure}

\begin{figure}[H]
\centering
\includegraphics[width=0.6\textwidth]{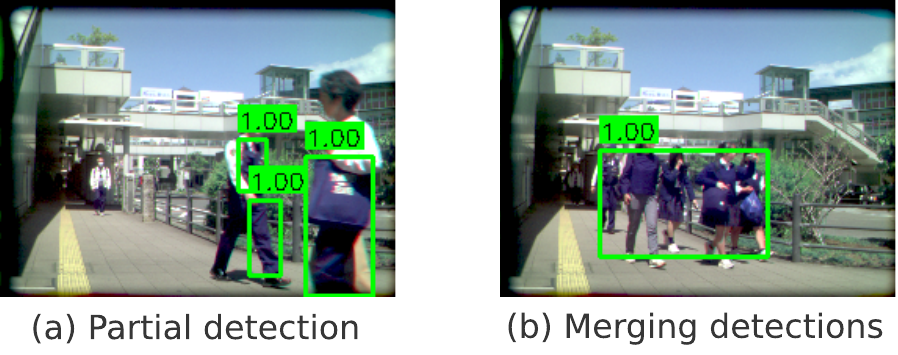}
\caption{Analysis of \hitomi{}'s failure cases for General Scene.
This figure illustrates typical failure modes of \hitomi{} in crowded public scenes. (a) Partial detection: Due to spectral limitations, only the upper body clothing is detected, resulting in a bounding box that is too small. (b) Merging detection: The algorithm incorrectly merges five closely located individuals into a single detection, failing to distinguish them as separate entities. These examples clarify the reasons for the lower AP score in the General Scene scenario.}
\label{fig:supp_basic_verification_3} 
\end{figure}

\begin{figure}[H]
\centering
\includegraphics[width=0.8\textwidth]{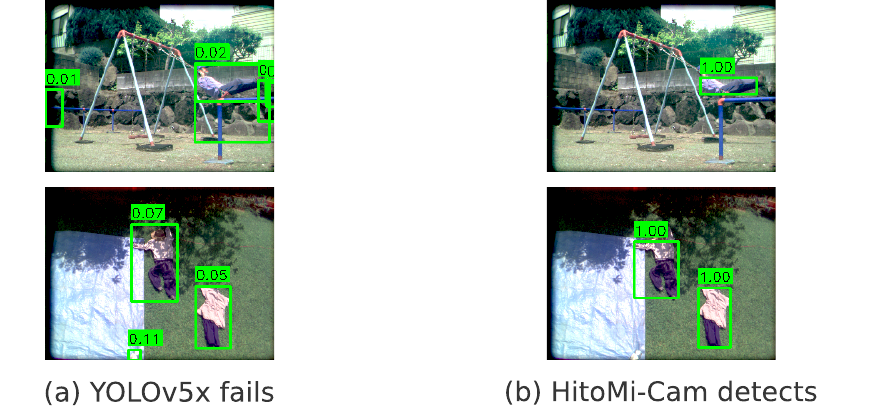}
\caption{Qualitative comparison with baseline detector (YOLOv5x) in challenging cases.
While the baseline method (a) fails, \hitomi{} (b) successfully detects the target. These examples highlight the robustness of \hitomi{} against non-human shapes (from Simulated SAR Scene) and significant motion (from Swing Scene).}
\label{fig:supp_basic_verification_4} 
\end{figure}

\begin{figure}[H]
\centering
\includegraphics[width=1.0\textwidth]{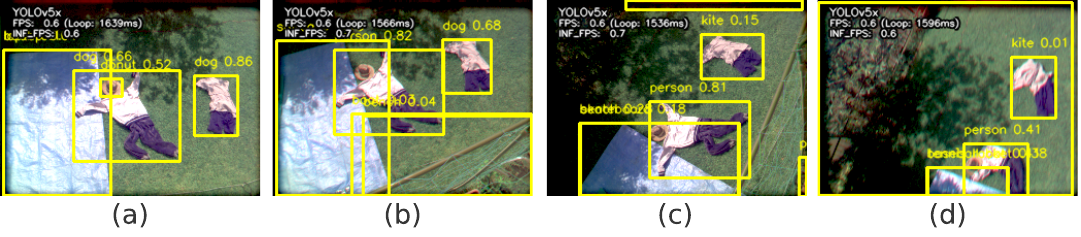}
\caption{Failure mode analysis of YOLOv5x on the Simulated SAR Scene.
Detection results with all 80 MS COCO classes enabled, illustrating the root cause of low person-class AP reported in Section 3.5.2. The smaller target is predominantly misclassified as "dog" with high confidence (a, b), while the larger simulated person is sporadically detected as "person" with varying confidence (c, d). This behavior demonstrates the fundamental challenge for shape-based detectors when classifying atypical, non-human-like shapes as the person class—targets that represent significant deviations from typical postures in the MS COCO training data.}
\label{fig:supp_basic_verification_5} 
\end{figure}

\begin{figure}[H]
\centering
\includegraphics[width=1.0\textwidth]{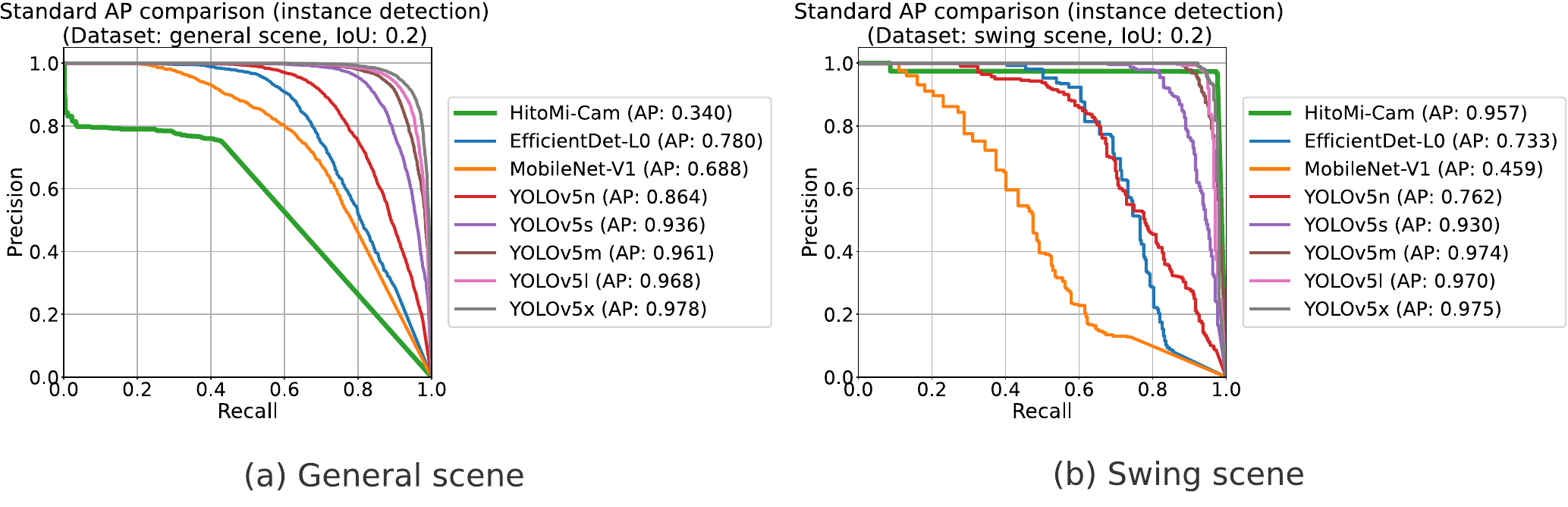}
\caption{PR curves of each model for General Scene and Swing Scene (IoU $\geq$ 0.2). This figure shows the PR curves of all compared models (\hitomi{}, YOLOv5 family, MobileNet-V1, and EfficientDet-L0). (a) For General Scene, the CNN-based models (especially the YOLOv5 family) significantly outperform \hitomi{}. (b) For Swing Scene, \hitomi{} achieves a high average precision (AP) and shows superiority over lightweight CNNs (such as YOLOv5n). The horizontal axis is recall, and the vertical axis is precision; the upper right quadrant shows higher performance.}
\label{fig:supp_pr_curves}
\end{figure}

\begin{figure}[H]
\centering
\includegraphics[width=1.0\textwidth]{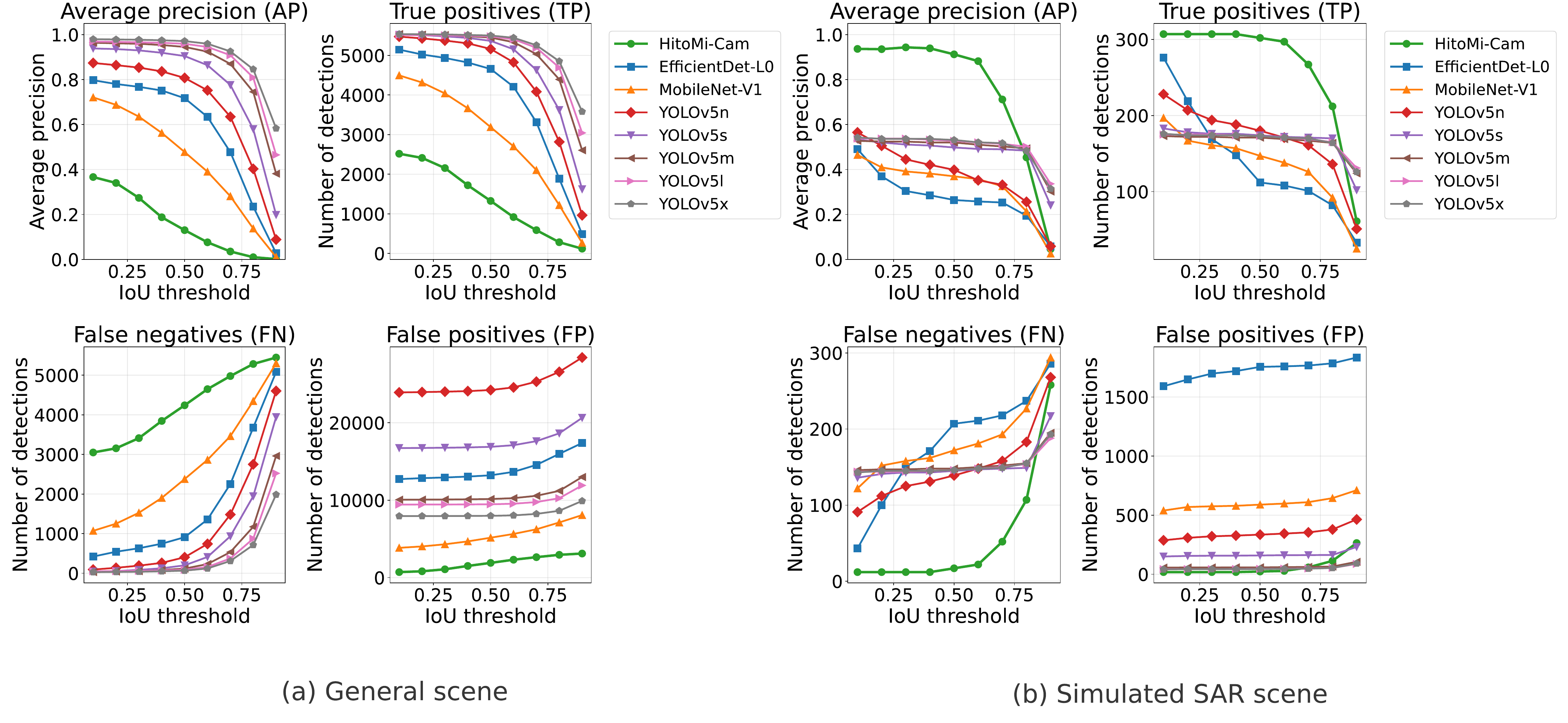}
\caption{Transition of performance metrics with respect to IoU threshold changes in (a) General Scene and (b) Simulated SAR Scene. Average Precision (AP), True Positives (TP), False Negatives (FN), and False Positives (FP) are shown. The sharp drop in the TP rate for \hitomi{}, which was prominent for Swing Scene, is not clear in the other scenarios, and high performance is maintained even at IoU = 0.5, especially for Simulated SAR Scene.}
\label{fig:supp_iou_analysis}
\end{figure}

\begin{figure}[H]
\centering
\includegraphics[width=0.8\textwidth]{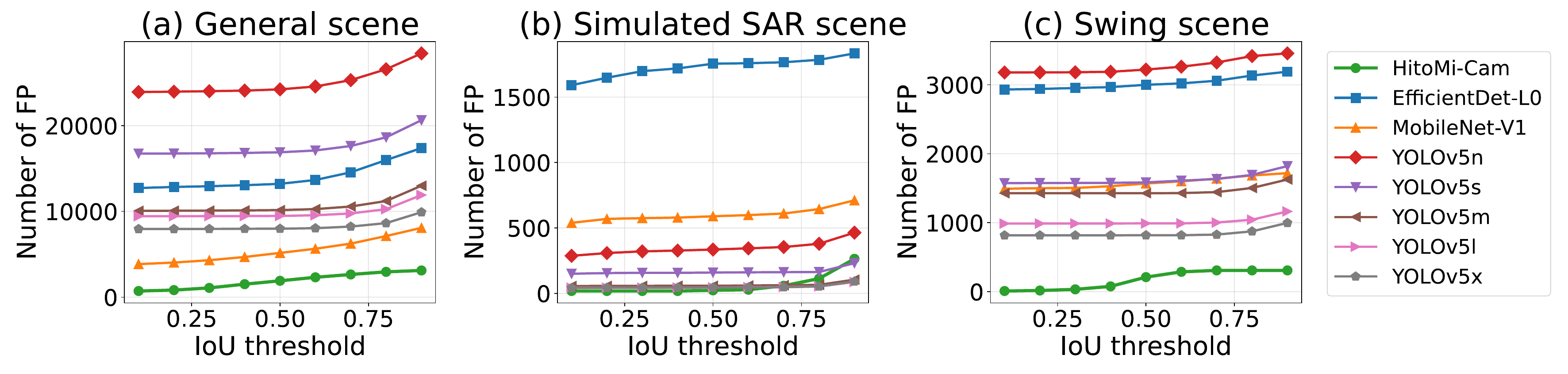}
\caption{Number of False Positives (FP) versus the IoU threshold across all three scenarios. This figure plots the number of FPs for \hitomi{} and the compared CNN models (YOLOv5 family, MobileNet-V1, and EfficientDet-L0) as a function of the IoU threshold: (a) General Scene, (b) Simulated SAR Scene, and (c) Swing Scene. Because \hitomi{} is designed to output only candidates with a confidence of 1.0 based on physical detection, it does not generate numerous low-confidence candidates like the CNN models do. Consequently, it exhibits a consistently lower number of FPs in all scenarios.}
\label{fig:supp_fp_analysis}
\end{figure}

\section*{Supplementary Videos}

\subsection*{Video S-1: Real-time operation demonstration on Raspberry Pi 5}

This video demonstrates the real-time operation of four detection methods running on the Raspberry Pi 5 single-board computer: \hitomi{}, YOLOv5n, MobileNet-V1, and EfficientDet-L0. The video shows a parallel comparison in a challenging dynamic scene where the target undergoes extreme posture changes.

Key observations from the video:
\begin{itemize}
    \item \hitomi{} operates at 23.2 fps and maintains stable detection throughout the sequence, continuously tracking the target despite rapid motion and atypical postures.
    \item YOLOv5n operates at 4.7 fps and shows intermittent detection instability for non-standard poses, despite its higher baseline accuracy in typical scenarios.
    \item MobileNet-V1 operates at 15.4 fps but exhibits occasional detection failures during extreme posture changes.
    \item EfficientDet-L0 operates at 7.6 fps and demonstrates moderate performance for this challenging scenario.
\end{itemize}

The video visually confirms that \hitomi{} achieves both high-speed processing and stable shape-agnostic detection on resource-constrained edge hardware, demonstrating its practical viability for real-time applications.

\noindent\textbf{Video specifications:}
\begin{itemize}
    \item Format: MP4 (H.264)
    \item Resolution: 546 $\times$ 480 pixels
    \item Duration: 12 seconds
    \item Availability: Included as an ancillary file (Video\_S-1.mp4)
\end{itemize}

\section*{Supplementary Tables}

\begin{tabularx}{\textwidth}{p{4.2cm} *{5}{>{\centering\arraybackslash}X}}

\caption{Comparison of processing times across different sensor modes and resolutions. This table presents the detailed processing time data (mean $\pm$ standard deviation) from the preliminary experiments that informed the selection of the operational configuration used in this study. The results confirm that a resolution of 253 $\times$ 190 pixels captured in binning mode (corresponding to "2K\_2" in the table) was the optimal configuration capable of achieving near-real-time performance (approx. 23.2 frames per second (fps); 43.1 ms total) on the Raspberry Pi 5 single-board computer (Raspberry Pi Ltd., Cambridge, UK). Consequently, this configuration was adopted for all subsequent evaluation experiments.}
\label{tab:performance_metrics_comparison} \\
\toprule
\textbf{Metric} & \textbf{4K\_1} & \textbf{4K\_2} & \textbf{4K\_4} & \textbf{2K\_1} & \textbf{2K\_2} \\
\midrule
\endfirsthead

\multicolumn{6}{c}{{\tablename\ \thetable{} -- continued from previous page}} \\
\toprule
\textbf{Metric} & \textbf{4K\_1} & \textbf{4K\_2} & \textbf{4K\_4} & \textbf{2K\_1} & \textbf{2K\_2} \\
\midrule
\endhead

\midrule \multicolumn{6}{r}{{Continued on next page}} \\
\endfoot

\bottomrule
\endlastfoot

Image resolution (pixels) & 1014 $\times$ 760 & 507 $\times$ 380 & 253 $\times$ 190 & 507 $\times$ 380 & 253 $\times$ 190 \\
\midrule
\multicolumn{6}{l}{\textit{Pre-processing (ms)}} \\
\quad Image acquisition & 0.0 $\pm$ 0.0 & 0.0 $\pm$ 0.0 & 40.8 $\pm$ 11.3 & 0.0 $\pm$ 0.0 & 0.0 $\pm$ 0.0 \\
\quad RAW data conversion & 10.9 $\pm$ 2.2 & 6.0 $\pm$ 0.4 & 6.1 $\pm$ 0.2 & 2.7 $\pm$ 0.7 & 2.3 $\pm$ 0.5 \\
\quad RAW reconstruction & 23.0 $\pm$ 3.2 & 19.0 $\pm$ 2.9 & 16.8 $\pm$ 0.5 & 6.5 $\pm$ 1.4 & 6.0 $\pm$ 1.7 \\
\quad Bayer processing & 0.0 $\pm$ 0.0 & 13.9 $\pm$ 0.6 & 3.7 $\pm$ 0.3 & 0.0 $\pm$ 0.0 & 3.8 $\pm$ 0.5 \\
\quad Luminance calculation & 26.8 $\pm$ 3.7 & 5.9 $\pm$ 0.7 & 1.4 $\pm$ 0.1 & 11.4 $\pm$ 1.5 & 2.0 $\pm$ 0.7 \\
\quad Luminance reconstruction & 29.9 $\pm$ 2.9 & 5.1 $\pm$ 0.9 & 1.5 $\pm$ 0.2 & 7.8 $\pm$ 1.2 & 2.5 $\pm$ 0.9 \\
Subtotal: Pre-processing (ms) & 115.8 $\pm$ 16.8 & 55.0 $\pm$ 6.0 & 30.1 $\pm$ 1.3 & 35.2 $\pm$ 5.7 & 17.8 $\pm$ 4.8 \\
\midrule
\multicolumn{6}{l}{\textit{MLP Inference (ms)}} \\
\quad Data type conversion & 6.9 $\pm$ 1.8 & 1.1 $\pm$ 0.2 & 0.3 $\pm$ 0.0 & 2.0 $\pm$ 0.4 & 0.5 $\pm$ 0.3 \\
\quad Data preparation & 18.4 $\pm$ 3.0 & 3.9 $\pm$ 0.4 & 0.5 $\pm$ 0.0 & 4.9 $\pm$ 0.5 & 0.7 $\pm$ 0.3 \\
\quad Inference (Core) & 173.6 $\pm$ 5.8 & 52.0 $\pm$ 5.1 & 8.6 $\pm$ 1.0 & 42.7 $\pm$ 1.3 & 9.8 $\pm$ 1.5 \\
Subtotal: MLP Inference (ms) & 228.0 $\pm$ 5.9 & 65.8 $\pm$ 5.2 & 11.9 $\pm$ 1.2 & 56.2 $\pm$ 1.3 & 13.4 $\pm$ 1.7 \\
\midrule
\multicolumn{6}{l}{\textit{Post-processing (ms)}} \\
\quad Pseudo-color generation & 73.6 $\pm$ 2.6 & 14.5 $\pm$ 1.1 & 3.0 $\pm$ 0.3 & 17.9 $\pm$ 1.7 & 4.4 $\pm$ 1.4 \\
\quad Result colorization & 13.6 $\pm$ 1.5 & 4.4 $\pm$ 0.7 & 0.7 $\pm$ 0.1 & 3.1 $\pm$ 0.3 & 0.8 $\pm$ 0.1 \\
\quad Bounding box generation & 86.7 $\pm$ 1.7 & 21.6 $\pm$ 0.7 & 5.3 $\pm$ 0.5 & 21.7 $\pm$ 0.7 & 5.5 $\pm$ 0.4 \\
Subtotal: Post-processing (ms) & 173.9 $\pm$ 5.9 & 40.5 $\pm$ 2.4 & 9.0 $\pm$ 0.8 & 42.8 $\pm$ 2.7 & 10.8 $\pm$ 1.9 \\
\midrule
Total Time per Frame (ms) & 520.2 $\pm$ 9.5 & 182.1 $\pm$ 7.9 & 179.4 $\pm$ 26.4 & 135.2 $\pm$ 2.5 & 43.1 $\pm$ 3.1 \\
\end{tabularx}
\keepXColumns

\begin{longtable}{p{0.5\textwidth}p{0.15\textwidth}p{0.25\textwidth}}
\caption{List of 41 labels used for MLP training. The table lists the 39 clothing categories and 2 background categories used to train the classifier. The clothing samples consist of common materials such as polyester, cotton, and wool and cover a wide range of colors. This table is adapted from the author's previous work.}
\label{tab:mlp_labels} \\
\toprule
\textbf{Label Name \textsuperscript{1}} & \textbf{Index} & \textbf{R-channel Intensity} \\
\midrule
\endfirsthead

\multicolumn{3}{c}{{\tablename\ \thetable{} -- continued from previous page}} \\
\toprule
\textbf{Label Name \textsuperscript{1}} & \textbf{Index} & \textbf{R-channel intensity} \\
\midrule
\endhead

\midrule \multicolumn{3}{r}{{Continued on next page}} \\
\endfoot

\bottomrule
\endlastfoot

\begin{filecontents*}{Train_Samples_ESC_Period.csv}
Label,Index,R-channel
Inorganic background,0,0
Plant background,34,255
P.Fluorescent-Yellow,1,53
P.Gray 2,2,245
P.White  + C.White,3,67
P.Black + C.Black,4,204
P.Gray 1 + W.Gray 1 + C.Lt. Gray,5,145
P.Fluorescent-Orange,6,135
P.Blue + C.Blue,7,219
P.Yellow + C.Yellow,8,127
P.Navy 2,9,187
P.Navy 1 + C.Navy,10,72
P.Red + C.Red + W.Red,11,178
P.Green + C.Green,12,248
C.Gray 2,13,244
C.Gray 1,14,189
C.Purple + W.Navy,15,64
W.Black + W.Jet-Black,16,197
W.Yellow,17,128
W.Gray 2,18,65
W.Khaki + C.Lt. Olive,19,171
W.Blue,20,241
W.White,21,153
W.Lt. Blue + P.Lt. Blue + C.Lt. Blue,22,240
W.Green,23,188
P.Sax Blue,25,246
C.Lavender,26,236
P.Lt. Green + C.Bright Green,27,252
C.Light Beige + W.J. Sand Beige,28,233
W.G. 1 Beige + W.T. 2 Beige,29,227
W.T. 1 Ivory Beige,30,214
P.Lt. Pink + C.Baby Pink + C.Lt. Pink,35,251
W.G. 13 Cream Yellow,36,206
P.MC. Pink,38,253
P.Lt. Yellow + C.Lt. Yellow,39,243
P.MC. Red,40,239
A. Orange + Glim. Orange,41,203
P.MC. Purple,43,247
W.J. 6 Magenta,45,221
C.Lt. Purple,46,232
P.Lt. Purple,47,249
\end{filecontents*}
\csvreader[
late after line=\\ \midrule,
]{Train_Samples_ESC_Period.csv}
{1=\Label,2=\Index,3=\Rchannel}{
\Label & \Index & \Rchannel
}
\end{longtable}

\noindent{\footnotesize{
\textsuperscript{1} Key to abbreviations: P. = polyester, 
C. = cotton,
W. = wool,
T. = Toray (a Japanese synthetic fabric manufacturer), G. = gabardine,
MC. = mixed color,
J. = Josette (a linen-based fabric), A. = athletic,
Glim. = glimmer,
and Lt. = light.
}}

\begin{tabularx}{\textwidth}{l l *{7}{>{\centering\arraybackslash}X}}
\caption{Detailed performance metrics of each model across all three scenarios with an IoU threshold of 0.2. This table shows all performance metrics (AP: Average Precision, Precision, Recall, F1 score, TP: True Positives, FP: False Positives, and FN: False Negatives) for all compared models in all three scenarios.}
\label{tab:supp_all_scenarios_iou02} \\
\toprule
\textbf{Scene} & \textbf{Model} & \textbf{AP} & \textbf{Prec.} & \textbf{Recall} & \textbf{F1} & \textbf{TP} & \textbf{FP} & \textbf{FN} \\
\midrule
\endfirsthead

\multicolumn{9}{c}{{\tablename\ \thetable{} -- continued from previous page}} \\
\toprule
\textbf{Scene} & \textbf{Model} & \textbf{AP} & \textbf{Prec.} & \textbf{Recall} & \textbf{F1} & \textbf{TP} & \textbf{FP} & \textbf{FN} \\
\midrule
\endhead

\midrule \multicolumn{9}{r}{{Continued on next page}} \\
\endfoot

\bottomrule
\endlastfoot

\multirow{8}{*}{\shortstack[l]{General\\Scene}} & HitoMi-Cam & 0.3401 & 0.7458 & 0.4333 & 0.6394 & 2412 & 822 & 3154 \\
 & EfficientDet-L0 & 0.7803 & 0.2810 & 0.9024 & 0.5901 & 5023 & 12850 & 543 \\
 & MobileNet-V1 & 0.6876 & 0.5169 & 0.7752 & 0.7706 & 4315 & 4033 & 1251 \\
 & YOLOv5n & 0.8641 & 0.1847 & 0.9754 & 0.4642 & 5429 & 23964 & 137 \\
 & YOLOv5s & 0.9356 & 0.2476 & 0.9903 & 0.5520 & 5512 & 16748 & 54 \\
 & YOLOv5m & 0.9612 & 0.3540 & 0.9923 & 0.6698 & 5523 & 10079 & 43 \\
 & YOLOv5l & 0.9680 & 0.3686 & 0.9903 & 0.6545 & 5512 & 9442 & 54 \\
 & YOLOv5x & 0.9784 & 0.4102 & 0.9941 & 0.6980 & 5533 & 7954 & 33 \\
\midrule
\multirow{8}{*}{\shortstack[l]{Simulated\\SAR Scene}} & HitoMi-Cam & 0.9353 & 0.9446 & 0.9624 & 0.9710 & 307 & 18 & 12 \\
 & EfficientDet-L0 & 0.3697 & 0.1173 & 0.6865 & 0.2731 & 219 & 1648 & 100 \\
 & MobileNet-V1 & 0.4094 & 0.2269 & 0.5235 & 0.4338 & 167 & 569 & 152 \\
 & YOLOv5n & 0.5056 & 0.4019 & 0.6489 & 0.5866 & 207 & 308 & 112 \\
 & YOLOv5s & 0.5202 & 0.5345 & 0.5580 & 0.6244 & 178 & 155 & 141 \\
 & YOLOv5m & 0.5243 & 0.7511 & 0.5392 & 0.6705 & 172 & 57 & 147 \\
 & YOLOv5l & 0.5376 & 0.8028 & 0.5486 & 0.6807 & 175 & 43 & 144 \\
 & YOLOv5x & 0.5363 & 0.8018 & 0.5455 & 0.6794 & 174 & 43 & 145 \\
\midrule
\multirow{8}{*}{\shortstack[l]{Swing\\Scene}} & HitoMi-Cam & 0.9574 & 0.9450 & 0.9766 & 0.9605 & 292 & 17 & 7 \\
 & EfficientDet-L0 & 0.7330 & 0.0801 & 0.8562 & 0.1952 & 256 & 2940 & 43 \\
 & MobileNet-V1 & 0.4589 & 0.1283 & 0.7391 & 0.2594 & 221 & 1501 & 78 \\
 & YOLOv5n & 0.7623 & 0.0838 & 0.9732 & 0.2061 & 291 & 3180 & 8 \\
 & YOLOv5s & 0.9301 & 0.1568 & 0.9799 & 0.3476 & 293 & 1576 & 6 \\
 & YOLOv5m & 0.9741 & 0.1722 & 0.9933 & 0.3620 & 297 & 1428 & 2 \\
 & YOLOv5l & 0.9702 & 0.2299 & 0.9866 & 0.4361 & 295 & 988 & 4 \\
 & YOLOv5x & 0.9755 & 0.2640 & 0.9799 & 0.4730 & 293 & 817 & 6 \\
\end{tabularx}

\begin{longtable}{lccccc}
\caption{Detailed performance metrics of each model in the Swing Scene. This table shows all the performance metrics (AP: Average Precision, TP: True Positives, FN: False Negatives, FP: False Positives) for all compared models in the Swing Scene, calculated by varying the IoU threshold in 0.1 increments.}
\label{tab:supp_swing_details} \\
\toprule
\textbf{Model} & \textbf{IoU Threshold} & \textbf{AP} & \textbf{TP} & \textbf{FN} & \textbf{FP} \\
\midrule
\endfirsthead

\multicolumn{6}{c}{{\tablename\ \thetable{} -- continued from previous page}} \\
\toprule
\textbf{Model} & \textbf{IoU Threshold} & \textbf{AP} & \textbf{TP} & \textbf{FN} & \textbf{FP} \\
\midrule
\endhead

\midrule \multicolumn{6}{r}{{Continued on next page}} \\
\endfoot

\bottomrule
\endlastfoot

\multirow{9}{*}{HitoMi-Cam} & 0.1 & 0.9991 & 299 & 0 & 10 \\
& 0.2 & 0.9574 & 292 & 7 & 17 \\
& 0.3 & 0.8736 & 275 & 24 & 34 \\
& 0.4 & 0.6048 & 232 & 67 & 77 \\
& 0.5 & 0.1196 & 98 & 201 & 211 \\
& 0.6 & 0.0063 & 21 & 278 & 288 \\
& 0.7 & 0 & 0 & 299 & 309 \\
& 0.8 & 0 & 0 & 299 & 309 \\
& 0.9 & 0 & 0 & 299 & 309 \\
\midrule
\multirow{9}{*}{EfficientDet-L0} & 0.1 & 0.7405 & 264 & 35 & 2932 \\
& 0.2 & 0.7330 & 256 & 43 & 2940 \\
& 0.3 & 0.6827 & 242 & 57 & 2954 \\
& 0.4 & 0.5176 & 229 & 70 & 2967 \\
& 0.5 & 0.4429 & 196 & 103 & 3000 \\
& 0.6 & 0.3835 & 175 & 124 & 3021 \\
& 0.7 & 0.2737 & 137 & 162 & 3059 \\
& 0.8 & 0.0640 & 61 & 238 & 3135 \\
& 0.9 & 0.0026 & 6 & 293 & 3190 \\
\midrule
\multirow{9}{*}{MobileNet-V1} & 0.1 & 0.4767 & 229 & 70 & 1493 \\
& 0.2 & 0.4589 & 221 & 78 & 1501 \\
& 0.3 & 0.4429 & 218 & 81 & 1504 \\
& 0.4 & 0.3164 & 192 & 107 & 1530 \\
& 0.5 & 0.1759 & 153 & 146 & 1569 \\
& 0.6 & 0.1041 & 122 & 177 & 1600 \\
& 0.7 & 0.0494 & 82 & 217 & 1640 \\
& 0.8 & 0.0074 & 37 & 262 & 1685 \\
& 0.9 & 0.0001 & 3 & 296 & 1719 \\
\midrule
\multirow{9}{*}{YOLOv5n} & 0.1 & 0.8124 & 292 & 7 & 3179 \\
& 0.2 & 0.7623 & 291 & 8 & 3180 \\
& 0.3 & 0.7057 & 289 & 10 & 3182 \\
& 0.4 & 0.4563 & 282 & 17 & 3189 \\
& 0.5 & 0.3351 & 250 & 49 & 3221 \\
& 0.6 & 0.2644 & 208 & 91 & 3263 \\
& 0.7 & 0.1816 & 147 & 152 & 3324 \\
& 0.8 & 0.0221 & 54 & 245 & 3417 \\
& 0.9 & 0.0015 & 15 & 284 & 3456 \\
\midrule
\multirow{9}{*}{YOLOv5s} & 0.1 & 0.9346 & 295 & 4 & 1574 \\
& 0.2 & 0.9301 & 293 & 6 & 1576 \\
& 0.3 & 0.9212 & 292 & 7 & 1577 \\
& 0.4 & 0.9015 & 291 & 8 & 1578 \\
& 0.5 & 0.8730 & 284 & 15 & 1585 \\
& 0.6 & 0.7946 & 259 & 40 & 1610 \\
& 0.7 & 0.7158 & 235 & 64 & 1634 \\
& 0.8 & 0.4355 & 175 & 124 & 1694 \\
& 0.9 & 0.0387 & 48 & 251 & 1821 \\
\midrule
\multirow{9}{*}{YOLOv5m} & 0.1 & 0.9745 & 297 & 2 & 1428 \\
& 0.2 & 0.9741 & 297 & 2 & 1428 \\
& 0.3 & 0.9740 & 297 & 2 & 1428 \\
& 0.4 & 0.9695 & 297 & 2 & 1428 \\
& 0.5 & 0.9694 & 297 & 2 & 1428 \\
& 0.6 & 0.9609 & 295 & 4 & 1430 \\
& 0.7 & 0.9127 & 281 & 18 & 1444 \\
& 0.8 & 0.6075 & 221 & 78 & 1504 \\
& 0.9 & 0.1728 & 98 & 201 & 1627 \\
\midrule
\multirow{9}{*}{YOLOv5l} & 0.1 & 0.9709 & 295 & 4 & 988 \\
& 0.2 & 0.9702 & 295 & 4 & 988 \\
& 0.3 & 0.9691 & 294 & 5 & 989 \\
& 0.4 & 0.9691 & 294 & 5 & 989 \\
& 0.5 & 0.9650 & 293 & 6 & 990 \\
& 0.6 & 0.9609 & 292 & 7 & 991 \\
& 0.7 & 0.9163 & 280 & 19 & 1003 \\
& 0.8 & 0.6931 & 239 & 60 & 1044 \\
& 0.9 & 0.2149 & 119 & 180 & 1164 \\
\midrule
\multirow{9}{*}{YOLOv5x} & 0.1 & 0.9755 & 293 & 6 & 817 \\
& 0.2 & 0.9755 & 293 & 6 & 817 \\
& 0.3 & 0.9754 & 293 & 6 & 817 \\
& 0.4 & 0.9752 & 293 & 6 & 817 \\
& 0.5 & 0.9695 & 291 & 8 & 819 \\
& 0.6 & 0.9684 & 291 & 8 & 819 \\
& 0.7 & 0.9284 & 282 & 17 & 828 \\
& 0.8 & 0.6952 & 235 & 64 & 875 \\
& 0.9 & 0.1913 & 112 & 187 & 998 \\
\end{longtable}

\end{document}